\let\mypdfximage\pdfximage
\def\pdfximage{\immediate\mypdfximage}

\documentclass[letterpaper, 11pt]{article}

\usepackage{algorithm}
\usepackage{algorithmic}
\usepackage{amsmath}
\usepackage{amssymb}
\usepackage{amsthm}
\usepackage[font=small]{caption} \usepackage{epsfig}
\usepackage{geometry}
\usepackage{graphicx}
\usepackage[colorlinks=true, citecolor=blue, filecolor=black, linkcolor=red, urlcolor=blue]{hyperref}
\usepackage{soul} \setstcolor{red}
\setulcolor{blue}

\newtheorem{theorem}{Theorem}

\usepackage[capitalise,nameinlink,noabbrev]{cleveref} 
\renewcommand{\[}{\left[}
\renewcommand{\]}{\right]}
\renewcommand{\(}{\left(}
\renewcommand{\)}{\right)}

\newcommand{\vvvert}{|\kern-1pt|\kern-1pt|}

\newcommand{\EE}{\mathbb{E}}

\newcommand{\RR}{\mathbb{R}}

\newcommand{\CD}{\mathcal{D}}

\newcommand{\CF}{\mathcal{F}}

\newcommand{\CL}{\mathcal{L}}

\newcommand{\CN}{\mathcal{N}}

\newcommand{\CU}{\mathcal{U}}

\newcommand{\CX}{\mathcal{X}}

\newcommand{\argmax}{\operatornamewithlimits{arg\,max}}

\newcommand{\DKL}{D_{\mathrm{KL}}}

\crefname{section}{Sec.}{Sec.}
\Crefname{section}{Section}{Sections}
\crefname{subsection}{Sec.}{Sec.}
\Crefname{subsection}{Section}{Sections}
\crefname{figure}{Fig.}{Fig.}
\Crefname{figure}{Figure}{Figures}
\crefname{equation}{Eqn.}{Eqn.}
\Crefname{equation}{Equation}{Equations}

\crefdefaultlabelformat{#2\textup{#1}#3}

\usepackage[caption=false]{subfig}

\title{Bayesian Sequential Optimal Experimental Design for\\Nonlinear Models Using Policy Gradient Reinforcement Learning}
\author{Wanggang Shen\footnote{Corresponding author:
    \href{mailto:wgshen@umich.edu}{wgshen@umich.edu}, University of Michigan, Ann Arbor, MI 48109, USA.}, Xun Huan\footnote{\href{mailto:xhuan@umich.edu}{xhuan@umich.edu}, University of Michigan, Ann Arbor, MI 48109, USA. \href{https://uq.engin.umich.edu}{https://uq.engin.umich.edu}}}

\date{}

\begin{document}

\maketitle

\begin{abstract}
We present a mathematical framework and computational methods to optimally design a finite number of sequential experiments. We formulate this sequential optimal experimental design (sOED) problem as a finite-horizon partially observable Markov decision process (POMDP) in a Bayesian setting and with information-theoretic utilities.
It is built to accommodate continuous random variables, general non-Gaussian posteriors, and expensive nonlinear forward models. sOED then seeks an optimal design policy that incorporates elements of both feedback and lookahead, generalizing the suboptimal batch and greedy designs. 
We solve for the sOED policy numerically via policy gradient (PG) methods from reinforcement learning, and derive and prove the PG expression for sOED. Adopting an actor-critic approach, we parameterize the policy and value functions using deep neural networks and improve them using gradient estimates produced from simulated episodes of designs and observations. The overall PG-sOED method is validated on a linear-Gaussian benchmark, and its advantages over batch and greedy designs are demonstrated through a 
contaminant source inversion problem in a convection-diffusion field.  
\end{abstract}

\section{Introduction}
\label{sec:intro}

Experiments are indispensable for scientific research. Carefully designed experiments can provide substantial savings for these often expensive data-acquisition opportunities. However,  designs based on heuristics are usually not optimal, especially for complex systems with high dimensionality, nonlinear responses and dynamics, multiphysics, and uncertain and noisy environments. 
Optimal experimental design (OED), while leveraging a
criteria based on a forward model that simulates the experiment process, systematically quantifies and maximizes the value of experiments.

OED for linear models 
\cite{Fedorov1972,Atkinson2007} 
uses criteria based on the information matrix derived from the model, which can be calculated analytically. Different operations on this matrix form the core of the well-known alphabetical designs, such as the $A$- (trace), $D$- (determinant), and $E$-optimal (largest eigenvalue) designs. 
Bayesian OED further incorporates the notion of prior and posterior distributions that reflect the uncertainty update as a result of the experiment data 
\cite{Berger1985, Chaloner1995}.
In particular, the Bayesian $D$-optimal criterion generalizes to the nonlinear setting under an information-theoretic perspective \cite{Lindley1956}, and is equivalent to the expected Kullback–Leibler (KL) divergence from the prior to the posterior. 
However, these OED criteria are generally intractable to compute for nonlinear models
and must be
approximated~\cite{Box1959,Ford1989,Chaloner1995,Muller2005,Ryan2016}. With advances in computing power and a need to tackle bigger and more complex systems in engineering and science, there is a growing interest, urgency, and opportunity for computational development 
of nonlinear OED methods~\cite{Ryan2003,Terejanu2012,Huan2013,Long2015,Weaver2016,Alexanderian2016,Tsilifis2017,Overstall2017,Beck2018,Kleinegesse2019,Foster2019,Wu2020}.

When designing multiple experiments, commonly used 
approaches are often suboptimal. The first is \emph{batch} (or static) design: it rigidly designs all experiments together \emph{a priori} using the aforementioned linear or nonlinear OED method, and does not offer any opportunity to adapt when new information becomes available (i.e. no feedback). 
The second is \emph{greedy} (or myopic) design 
\cite{Box1992, Dror2008, Cavagnaro2010, Solonen2012, Drovandi2013, Drovandi2014, Kim2014,Hainy2016,Kleinegesse2021}:
it plans only for the \emph{next} experiment, 
updates with its observation, and repeats the design process. While greedy design has feedback, it lacks consideration for future effects and consequences (i.e. no lookahead). Hence, greedy design does not see the big picture or plan for the future. It is easy to relate, even from everyday experience (e.g., driving a car, planning a project), that a lack of feedback (for adaptation) and lookahead (for foresight) can lead to suboptimal decision-making with 
undesirable  consequences.

A provably optimal formulation of sequential experimental design---we refer to as sequential OED (sOED)~\cite{Muller2007,VonToussaint2011,Huan2015,Huan2016}---needs both elements of feedback and lookahead, and 
generalizes the batch and greedy designs. The main features
of sOED are twofold. First, sOED works with design \emph{policies} (i.e. functions that can adaptively suggest what experiment to perform depending on the current situation) in contrast to
static design values. Second, sOED always designs for all remaining experiments, thus capturing the effect on the entire future horizon when each design decision is made. 
Formally, the sOED problem can be formulated as a {partially observable Markov decision process} (POMDP). Under this agent-based view, the experimenter (agent) selects the experimental design (action) following a policy, and observes the experiment measurements (observation) in order to maximize the total utility (reward) that depends on the unknown model parameters (hidden state). 
A belief state 
can be further formed
based on the Bayesian posterior that describes the uncertainty 
of the hidden state, thereby turning the POMDP into a 
belief Markov decision process (MDP) \cite{littman1995learning}. 

The sOED problem targeted in our paper presents an atypical and challenging POMDP: finite horizon, continuous random variables, uncountably infinite belief state space, deterministic policy, continuous designs and observations, sampling-only transitions that each involves a Bayesian inference, and information measures as rewards. Thus, while there exists an extensive POMDP literature (e.g.,~\cite{cassandra1994acting, littman1995efficient, cassandra1998survey, kurniawati2016online, igl2018deep}), off-the-shelf methods cannot be directly applied to this sOED problem. 
At the same time, attempts for sOED have been sparse, with examples~\cite{Carlin1998,Gautier2000,Pronzato2002, Brockwell2003, Christen2003, Murphy2003, Wathen2006} 
focusing on discrete settings 
and with special problem and solution forms,
and do not use an information criteria or do not adopt a Bayesian framework. 
More recent efforts for Bayesian sOED~\cite{Huan2015,Huan2016} employ approximate dynamic programming (ADP) and transport maps, and illustrate the advantages of sOED over batch and greedy designs. However, this ADP-sOED method remains computationally expensive.

In this paper, we create new methods to solve the sOED problem in a computationally efficient manner, by drawing the state-of-the-art from reinforcement learning (RL) \cite{watkins1992q, sutton2000policy, szepesvari2010algorithms, mnih2015human, schulman2015trust, silver2016mastering, silver2017mastering, li2017deep, sutton2018reinforcement}.
RL approaches are often categorized as value-based (learn value functions only)
\cite{watkins1992q,mnih2015human,wang2016dueling,van2016deep}, policy-based (learn policy only)
\cite{willianms1988toward, williams1992simple}, or actor-critic (learn policy and value functions together) \cite{konda2000actor, peters2008natural, silver2014deterministic, lillicrap2015continuous}. 
ADP-sOED~\cite{Huan2015,Huan2016} is thus value-based, where the policy is only implicitly expressed via the learnt value functions. Consequently, each policy evaluation involves optimizing the value functions on-the-fly, a costly calculation especially for continuous action spaces. 
Both policy-based and actor-critic methods are more efficient in this respect. 
Actor-critic methods have further been observed to produce lower solution variance and faster convergence \cite{sutton2018reinforcement}. 

We adopt an actor-critic approach in this work. 
Representing and learning the policy explicitly further enables the use of policy gradient (PG) techniques \cite{sutton2000policy, kakade2001natural, degris2012off, silver2014deterministic, lillicrap2015continuous, schulman2015trust, mnih2016asynchronous, schulman2017proximal, lowe2017multi, liu2017stein, barth2018distributed} that estimate the gradient with respect to policy parameters, and in turn permits the use of gradient-based optimization algorithms.
Inspired by deep deterministic policy gradient (DDPG)~\cite{lillicrap2015continuous}, we further employ deep neural networks (DNNs) to parameterize and approximate the policy and value functions. The use of DNNs can take advantage of potentially large number of episode samples generated from the transition simulations, and compute gradients efficiently through back-propagation. 
Nevertheless, care needs be taken to design the DNNs and their hyperparameters in order to 
obtain stable and rapid convergence to a good sOED policy, which we will describe in the paper. 

The main contributions of our paper are as follows.
\begin{itemize}
\item We formulate the sOED problem as a finite-horizon POMDP under a Bayesian setting for continuous random variables, and illustrate its generalization over the batch and greedy designs.
\item We present the PG-based sOED (that we call PG-sOED) algorithm, proving the key gradient expression and proposing its Monte Carlo estimator. We further present the DNN architectures for the policy and value functions, and detail the numerical setup of the overall method.
\item We demonstrate the speed and optimality advantages of PG-sOED over ADP-sOED, batch, and greedy designs, on a benchmark and a problem of contaminant source inversion in a convection-diffusion field that involves an expensive forward model. 
\item We make available our PG-sOED code at \url{https://github.com/wgshen/sOED}. 
\end{itemize}

This paper is organized as follows. \Cref{sec:formulation} introduces the components needed in an sOED problem, culminating with the sOED problem statement. \Cref{sec:method} describes the details of the entire PG-sOED method.
\Cref{sec:results} presents numerical examples, a linear-Gaussian benchmark and a problem of contaminant source inversion in a convection-diffusion field, to validate PG-sOED and demonstrate its advantages over other baselines.
Finally, \cref{sec:conclusions} concludes the paper and provides an outlook for future work.

\section{Problem Formulation}
\label{sec:formulation}

\subsection{The Bayesian Paradigm}

We consider designing a finite\footnote{In experimental design, the experiments are generally expensive and limited in number. Finite and small values of $N$ are therefore of interest. 
This is in contrast to RL that often deals with infinite horizon.} number of $N$ experiments, indexed by  integers $k=0,1,\ldots,N-1$.
While the decision of how many experiments to perform (i.e. choice of $N$) is important, it is 
not considered
in this paper; instead, we assume $N$ is given and fixed.
Furthermore, let 
$\theta\in \RR^{N_{\theta}}$ denote the unknown model parameter we seek to 
learn
from the experiments, $d_k \in \CD_k\subseteq \RR^{N_d}$ the experimental design variable for the $k$th experiment (e.g., 
experiment conditions),
$y_k \in \RR^{N_y}$ the noisy observation from the $k$th experiment (i.e. experiment measurements), and $N_{\theta}$, $N_{d}$, and $N_{y}$ respectively the dimensions of parameter, design, and observation spaces. We further consider continuous $\theta$, $d_k$, and $y_k$, although discrete or mixed settings can be accommodated as well.
For simplicity,
we also let $N_d$ and $N_y$ be constant across all experiments, but this is not a requirement.

A Bayesian approach treats $\theta$ as a random variable. 
After performing the $k$th 
experiment, its conditional probability density function (PDF) is described by Bayes' rule:
\begin{align}
    \label{eq:bayes_rule}
    p(\theta|d_k,y_k,I_k) = \frac{p(y_k|\theta,d_k,I_k)p(\theta|I_k)}{p(y_k|d_k,I_k)}
\end{align}
where $I_k=\{ d_0,y_0,\dots,d_{k-1},y_{k-1} \}$ (and $I_0=\emptyset$) is the information set collecting the design and observation records from all experiments prior to the $k$th experiment, $p(\theta|I_k)$ is the prior PDF for the $k$th experiment,
$p(y_k|\theta,d_k,I_k)$ is the likelihood function,
$p(y_k|d_k,I_k)$ is the model evidence (or marginal likelihood, which is constant with respect to $\theta$),
and $p(\theta|d_k,y_k,I_k)$ is the posterior PDF. The prior is then a representation of the uncertainty about $\theta$ before 
the $k$th experiment, and the posterior describes the updated uncertainty about $\theta$ after having observed the outcome from the $k$th experiment.
In \cref{eq:bayes_rule}, we also simplify the prior $p(\theta|d_k,I_k)=p(\theta|I_{k})$, invoking a reasonable assumption that knowing only the design for $k$th experiment (but without knowing its outcome) would not affect the prior. 
The likelihood function carries the relation between the hidden parameter $\theta$ and the observable $y_k$, through a forward model $G_k$ that governs the underlying process for the $k$th experiment (e.g., constrained via a system of partial differential equations (PDEs)). For example, a common likelihood form is 
\begin{align}
    y_k = G_k(\theta, d_k; I_k) + \epsilon_k,
\end{align}
where $\epsilon_k$ is a Gaussian random variable that describes the discrepancy between model prediction $G_k$ and observation $y_k$ due to, for instance, measurement noise. The inclusion of $I_k$ in $G_k$ signifies that model behavior may be affected by previous experiments. Each evaluation of the likelihood $p(y_k|\theta,d_k,I_k) = p_{\epsilon}(y_k-G_k(\theta,d_k; I_k))$ thus involves a forward model solve, typically the most expensive part of the computation.
Lastly, 
the posterior $p(\theta|d_k,y_k,I_k)=p(\theta|I_{k+1})$ becomes the prior for the $(k+1)$th experiment via the same form of \cref{eq:bayes_rule}. Hence, Bayes' rule can be consistently and recursively applied for a sequence of multiple experiments. 

\subsection{Sequential Optimal Experimental Design}
\label{sec:math_formulation}

\begin{figure}[htb]
  \centering
  \includegraphics[width=0.95\linewidth]{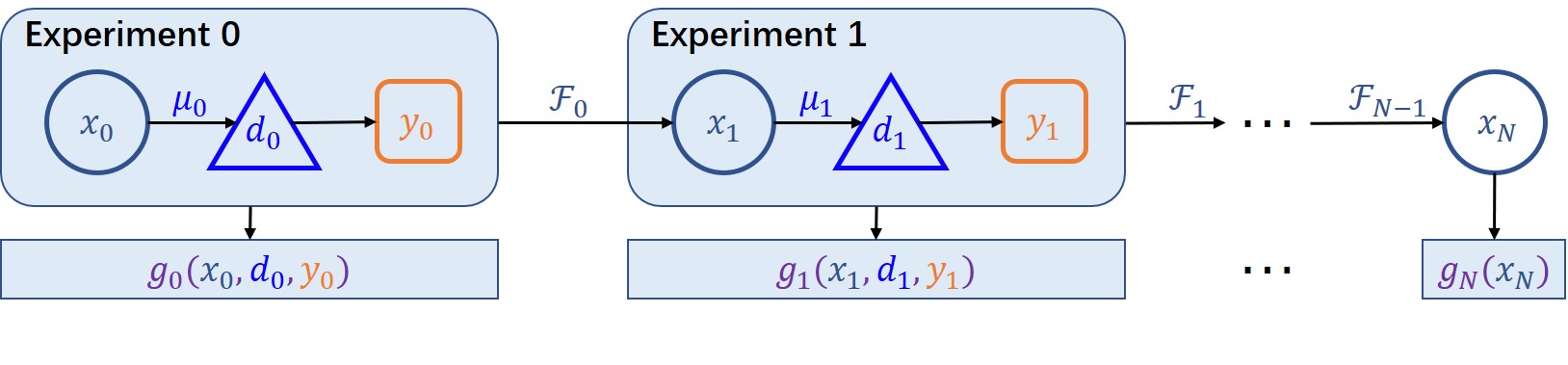}
  \caption{Flowchart of the process involved in a $N$-experiment sOED.}
  \label{fig:process}
\end{figure}

We now present a general framework for sOED, posed as a POMDP.
An overview flowchart for sOED is presented in \cref{fig:process} to accompany the definitions below.

\textbf{State.} We introduce the state variable $x_k=[x_{k,b},x_{k,p}] \in \CX_k$
to be the state prior to designing and performing the $k$th experiment. 
Hence, 
$x_0,\ldots,x_{N-1}$ denote the respective states prior to each of the $N$ experiments, and $x_1,\ldots,x_N$ denote the respective states after each of the $N$ experiments.
The state is an entity that summarizes past information needed for making experimental design decisions in the future. 
It is very general and can contain different quantities deemed to be  decision-relevant. 
In our case here, the state consists of a belief state $x_{k,b}$ reflecting our state of uncertainty about the hidden $\theta$, and a physical state $x_{k,p}$ carrying other non-random variables pertinent to the design problem. 
Since $\theta$ is not observable and can be only inferred from noisy and indirect observations $y_k$ through Bayes' rule in \cref{eq:bayes_rule}, this setup can be viewed as a POMDP for $\theta$ (or a MDP for $x_k$).

Conceptually, a \emph{realization} of the belief state manifests as 
the continuous posterior (conditional) random variable 
$(x_{k,b} = x'_{k,b}) = (\theta|I_k=I_k')$, 
where the prime denotes realization. Such a random variable can be 
portrayed by, for example, its PDF, cumulative distribution function, or characteristic function\footnote{
It is possible for $\theta|I_k$'s with different $I_k'$'s to have the same PDF (or distribution or characteristic function), for example simply by exchanging the experiments. Hence, the mappings from $I_k$ to these portrayals (PDF, distribution, characteristic functions) are non-injective. This may be problematic when considering transition probabilities of the belief state, but avoided if we keep to our root definition of belief state based on $I_k$, which remains unique.}.
Attempting to directly represent these infinite-dimensional quantities  in practice would require some finite-dimensional approximation or discretization.
Alternatively, one can adopt a nonparametric approach and track 
$I_k$
(from a given initial $x_0$),
which then yields a 
representation of $x_{k}$ (both $x_{k,b}$ and $x_{k,p}$) without any approximation\footnote{$I_k$ collects the complete history of experiments and their observations, therefore is a sufficient statistic for $x_k$ by definition. Hence, if $I_k$ is known, then the full state $x_k$ is equivalently represented. 
All of these are conditioned on a given initial $x_0$ (which includes the prior on $\theta$), but for simplicity we will omit this conditioning when writing the PDFs in this paper, with the understanding that it is always implied. }
but its dimension grows with $k$. However, the dimension is always bounded since the maximum number of experiments considered is finite (i.e. $k < N$).
In any case, the belief state space is uncountably infinite since $\theta$ is a continuous random variable (i.e. the possible posteriors that can be realized is uncountably infinite).
We will further detail our numerical representation of the belief state in \cref{sec:policy_net} and \cref{sec:numerical_belief_state}.

\textbf{Design (action) and policy.} Sequential experimental design involves building policies mapping from the state space to the design space, $\pi = \{\mu_k : \CX_k \mapsto \CD_k, k=0,\ldots,N-1
\}$, such that the design for the $k$th experiment is determined by the state via $d_k=\mu_k(x_k)$. Thus, sequential design is inherently adaptive, computing designs based on the current state which depends on the previous experiments and their outcomes.
We focus on deterministic policies in this study, where policy functions $\mu_k$ produce deterministic outputs. 

\textbf{System dynamics (transition function).} The system dynamics, denoted by $x_{k+1}=\CF_k(x_k,d_k,y_k)$, describes the transition from state $x_k$ to state $x_{k+1}$ after carrying out the $k$th experiment with design $d_k$ and observation $y_k$. For the belief state, the prior $x_{k,b}$ can be updated to the posterior $x_{k+1,b}$ via Bayes' rule in \cref{eq:bayes_rule}. The physical state, if present, evolves based on the relevant physical process.
While the system dynamics described in \cref{eq:bayes_rule} appears deterministic given a specific realization of $d_k$ and $y_k$, it is a stochastic transition since the observation $y_k$ is random. In particular, there exists an underlying transition probability
\begin{align}
p(x_{k+1}|x_{k},d_{k})=p(y_k|x_k,d_k)=p(I_{k+1}|d_{k},I_{k}) =p(y_{k}|d_{k},I_{k}) = 
\int_{\Theta} p(y_k|\theta,d_k, I_k)p(\theta|I_{k})\,d\theta,
\label{eq:transition}
\end{align}
where we 
simplify the prior with $p(\theta|d_k,I_k)=p(\theta|I_{k})$. 
This transition probability is intractable and does not have a closed form. However, we are able to generate samples of the next state by sampling from the prior and likelihood, as suggested by the last equality in \cref{eq:transition}. Hence, we have a model-based (via a sampling model) setup.

\textbf{Utility (reward).} We denote $g_k(x_k,d_k,y_k) \in \RR$ to be the immediate reward from performing an experiment. Most generally, this quantity can depend on the state, design, and observation. For example, it may simply be the (negative) cost of the $k$th experiment.
Similarly, we define a terminal reward $g_N(x_N) \in \RR$ containing any additional reward measure that reflects the benefit of reaching certain final state, and that can only be computed after the entire set of experiments is completed. We will provide a specific example of reward functions pertaining to information measures in \cref{sec:information_gain}.

\textbf{sOED problem statement.} The sOED problem seeks the policy that solves the following optimization problem: 
from a given initial state $x_0$, \begin{align}
    \label{eq:optimal_policy}
    \pi^\ast = \argmax_{\pi=\{\mu_0,\ldots,\mu_{N-1}\}}& \qquad U(\pi)\\
    \text{s.t.}& 
    \qquad d_k = \mu_k(x_k) \in \CD_k, \nonumber\\
    &\qquad x_{k+1}=\CF_k(x_k,d_k,y_k),
    \hspace{3em} \text{for}\quad k=0,\dots,N-1, \nonumber
\end{align}
where
\begin{align}
    \label{eq:expected_utility}
    U(\pi) = \EE_{y_0,...,y_{N-1}|\pi,x_0}\[\sum_{k=0}^{N-1}g_k(x_k,d_k,y_k)+g_N(x_N)\]
\end{align}
is the expected total utility functional. 
While here $x_0$ is fixed, this formulation can easily be adjusted to accommodate stochastic $x_0$ as well, by including $x_0$ as a part of $I_k$ and taking another expectation over $x_0$ in \cref{eq:expected_utility}. 

Overall, our sOED problem corresponds to a model-based planning problem of RL. It is challenging for several reasons: 
\begin{itemize}
\item finite horizon, where the policy functions $\mu_k$ are different for each $k$ and need to be tracked and solved for separately; 
\item partially and indirectly observed hidden $\theta$ whose belief state space is uncountably infinite and also infinite-dimensional or nonparametric; 
\item deterministic policy;
\item continuous design (action) and observation spaces; 
\item transition probability intractable to compute, and transition can only be sampled;
\item each belief state transition involves a Bayesian inference, 
requiring many 
forward model evaluations; 
\item reward functions are information measures for continuous random variables (discussed below), which are difficult to estimate.
\end{itemize}

\subsection{Information Measures as Experimental Design Rewards}
\label{sec:information_gain}

We wish to adopt reward functions 
that reflect the degree of success for the experiments, 
not only
the experiment costs. Determining such an appropriate quantity depends on the experimental goals, e.g., to achieve inference, prediction, model discrimination, etc. One popular choice corresponding to the goal of parameter inference is to maximize a measure of information gained on $\theta$. 
Lindley's seminal paper~\cite{Lindley1956} proposes to use the mutual information between the parameter and observation as the expected utility, and Ginebra~\cite{Ginebra2007} provides more general criteria for proper measure of information gained from an experiment. 
From the former, mutual information is equal to the expected KL
divergence from the prior to the posterior. The KL divergence provides an intuitive interpretation as it quantifies the farness between the prior and the posterior distributions, and thus a larger divergence corresponds to a greater degree of belief update---and hence information gain---resulting from the experiment and its observation.

In this paper, we follow Lindley's approach and demonstrate the use of KL divergence as sOED rewards,
and present two reasonable sequential design formulations that are in fact equivalent. The first, call it the \emph{terminal formulation}, involves clumping the information gain from all $N$ experiments in the terminal reward (for clarity, we omit all other reward contributions common to the two formulations, although it would be trivial to show the equivalence for those cases too): 
\begin{align}
 g_k(x_k, d_k, y_k) &= 0, \qquad k=0,\ldots,N-1 \label{eq:terminal1}\\
 g_N(x_N) &= \DKL\( p(\cdot|I_N)\,||\,p(\cdot|I_0) \) \nonumber\\ &= \int_{\Theta} p(\theta|I_N) \ln\[\frac{p(\theta|I_N)}{p(\theta|I_0)}\]\,d\theta.\label{eq:terminal_info_gN}
\end{align}
The second, call it the \emph{incremental formulation}, entails the use of incremental information gain from each experiment in their respective immediate rewards:
\begin{align}
    g_k(x_k, d_k, y_k) &= \DKL\( p(\cdot|I_{k+1})\,||\,p(\cdot|I_k) \) \nonumber\\&= \int_{\Theta} p(\theta|I_{k+1}) \ln\[\frac{p(\theta|I_{k+1})}{p(\theta|I_k)}\]\,d\theta, \qquad k=0,\ldots,N-1\label{eq:incremental1}\\
    g_N(x_N) &= 0. \label{eq:incremental2}
\end{align}

\begin{theorem} 
\label{prop:terminal_incremental}
Let $U_T(\pi)$ be the sOED expected utility defined in \cref{eq:expected_utility} subject to the constraints in \cref{eq:optimal_policy} for a given policy $\pi$ while using the terminal formulation \cref{eq:terminal1,eq:terminal_info_gN}. Let $U_I(\pi)$ be the same except using  the incremental formulation \cref{eq:incremental1,eq:incremental2}. Then $U_T(\pi)=U_I(\pi)$. 
\end{theorem}

A proof is provided in \cref{app:incre_terminal}.
As a result, the two formulations correspond to the same sOED problem. 

\subsection{Generalization of Suboptimal Experimental Design Strategies}
\label{sec:subopt_design}

We also make the connection between sOED to 
the commonly used batch design and greedy sequential design.
We illustrate below that both batch and greedy designs are,
in general, suboptimal with respect to the expected utility \cref{eq:expected_utility}. Thus, sOED generalizes these design strategies.

Batch OED designs all $N$ experiments together prior to performing any of those experiments. Consequently, it is non-adaptive, and cannot make use of  new information acquired from any of the $N$ experiments to help adjust the design of other experiments. Mathematically, batch design seeks static design values (instead of a policy) over the joint design space $\CD:=\CD_0 \times \CD_1 \times \cdots \times \CD_{N-1}$:
\begin{align}
    (d_0^{\mathrm{ba}},\dots,d_{N-1}^{\mathrm{ba}}) = \argmax_{(d_0,\dots,d_{N-1}) \in \CD} \EE_{y_0,\dots,y_{N-1}|d_0,\dots,d_{N-1},x_0}\[ \sum_{k=0}^{N-1}g_k(x_k,d_k,y_k) + g_N(x_N) \],\label{eq:batch}
\end{align}
subject to the system dynamics. In other words, the design $d_k$ is chosen independent of $x_k$ (for $k > 0$).
The suboptimality of batch design becomes clear once realizing \cref{eq:batch} is equivalent to the sOED formulation in \cref{eq:optimal_policy} but restricting all $\mu_k$ to be only constant functions. Thus, $U(\pi^{\ast}) \geq U(\pi^{\mathrm{ba}}=d^{\mathrm{ba}})$. 

Greedy design is also a type of sequential experimental design and produces a policy. It optimizes only for the immediate reward at each experiment:
\begin{align}
\mu_k^{\mathrm{gr}} = \argmax_{\mu_k} \EE_{y_k|x_k,\mu_k(x_k)}\[ g_k(x_k,\mu_k(x_k),y_k) \], \qquad k=0,\dots,N-1,\label{eq:greedy}
\end{align}
without needing to subject to the system dynamics since the policy functions $\mu_k^{\mathrm{gr}}$ are decoupled. $U(\pi^{\ast}) \geq U(\pi^{\mathrm{gr}})$ follows trivially.
As a more specific example when using information measure utilities described in \cref{sec:information_gain}, greedy design would only make sense under the incremental formulation (\cref{eq:incremental1,eq:incremental2}).
Then, together with \cref{prop:terminal_incremental}, we have 
$U_{T}(\pi^{\ast})=U_{I}(\pi^{\ast}) \geq U_{I}(\pi^{\mathrm{gr}})$.

\section{Policy Gradient for Sequential Optimal Experimental Design}
\label{sec:method}

We approach the sOED problem by directly parameterizing the policy functions and representing them explicitly. We then develop gradient expression with respect to the policy parameters, so to enable gradient-based optimization for numerically identifying optimal or near-optimal policies. Such approach is known as the PG
method (e.g., \cite{silver2014deterministic, lillicrap2015continuous}).
In addition to the policy, we also parameterize and learn the value functions,
thus arriving at an actor-critic form. 
PG contrasts with previous  ADP-sOED efforts~\cite{Huan2015,Huan2016} that 
approximate only the value functions. In those works, the policy is represented implicitly, and requires solving a (stochastic) optimization problem each time the policy is evaluated. This renders both the offline training and online policy usage computationally expensive. As we will demonstrate, PG sidesteps this requirement.

In the following, we first derive the exact PG expression in \cref{ss:PG_exact}. We then present numerical methods in \cref{ss:PG_numerical} to estimate this exact PG expression. In particular, this requires adopting a parameterization of the policy functions; we will present the use of DNNs to achieve this parameterization. Once the policy parameterization is established, we can then compute the PG with respect to the parameters, and optimize them using a gradient ascent procedure.

\subsection{Derivation of the Policy Gradient}
\label{ss:PG_exact}

The PG approach to sOED (PG-sOED) involves parameterizing each policy function $\mu_{k}$  with parameters $w_k$ ($k=0,\ldots,N-1$), which we denote by the shorthand form $\mu_{k,w_k}$. In turn, the policy $\pi$ is parameterized by $w=\{w_k, \forall k\} \in \RR^{N_w}$ 
and denoted by $\pi_{w}$, where $N_w$ is the dimension of the overall policy parameter vector. The sOED problem statement from \cref{eq:optimal_policy,eq:expected_utility} then updates to: from a given initial state $x_0$,
\begin{align}
    \label{eq:PG_sOED}
    w^{\ast} = \argmax_{w}& \qquad U(w)\\
    \text{s.t.}& 
    \qquad d_k = \mu_{k,w_k}(x_k) \in \CD_k, \nonumber\\
    &\qquad x_{k+1}=\CF_k(x_k,d_k,y_k), 
    \hspace{3em} \text{for}\quad k=0,\dots,N-1, \nonumber
\end{align}
where
\begin{align}
    \label{eq:expected_utility_w}
    U(w) = \EE_{y_0,...,y_{N-1}|\pi_w,x_0}\[\sum_{k=0}^{N-1}g_k(x_k,d_k,y_k)+g_N(x_N)\].
\end{align}
We now aim to derive the gradient $\nabla_{w} U(w)$.

Before presenting the gradient expression, we need to introduce the value functions. 
The \emph{state-value function} (or \emph{V-function}) following policy $\pi_{w}$ and at the $k$th experiment is
\begin{align}
V_k^{\pi_{w}}(x_k)&=\EE_{y_k,\dots,y_{N-1}|\pi_{w},x_k}\[\sum_{t=k}^{N-1} g_t(x_t,\mu_{t,w_t}(x_t),y_t) + g_N(x_N)\] \\
 &= \EE_{y_k|\pi_w,x_k} \[ g_k(x_k,\mu_{k,w_k}(x_k),y_k) + V^{\pi_w}_{k+1}(x_{k+1}) \] \\
 V_N^{\pi_{w}}(x_N) &= g_N(x_N)
\end{align}
for $k=0,\ldots,N-1$, where $x_{k+1}=\CF_k(x_k,\mu_{k,w_k}(x_k),y_k)$.
The V-function is the expected cumulative remaining reward starting from a given state $x_k$ and following policy $\pi_{w}$ for all remaining experiments. 
The \emph{action-value function} (or \emph{Q-function}) following policy $\pi_{w}$ and at the $k$th experiment is
\begin{align}
\label{eq:action_bellman}
Q_k^{\pi_{w}}(x_k,d_k)&=\EE_{y_k,\dots,y_{N-1}|\pi_{w},x_k,d_k}\[g_k(x_k,d_k,y_k) + \sum_{t=k+1}^{N-1} g_t(x_t,\mu_{t,w_t}(x_t),y_t) + g_N(x_N)\]
\\
&=\EE_{y_k|x_k,d_k} \[ g_k(x_k,d_k,y_k) + Q^{\pi_w}_{k+1}(x_{k+1},\mu_{k+1,w_{k+1}}(x_{k+1}))\]
\label{eq:action_bellman2}
\\
Q_{N}^{\pi_{w}}(x_N,\cdot) &= g_N(x_N).
\end{align}
for $k=0,\ldots,N-1$, where $x_{k+1}=\CF_k(x_k,d_k,y_k)$. 
The Q-function is the expected cumulative remaining reward for performing the $k$th experiment at the given design $d_k$ from a given state $x_k$ and thereafter following policy $\pi_{w}$. The two functions are related via
\begin{align}
V_k^{\pi_{w}}(x_k)=Q_k^{\pi_{w}}(x_k,\mu_{k,w_k}(x_k)).
\end{align}

\begin{theorem}
\label{thm:PG}
The gradient of the expected utility in \cref{eq:expected_utility_w} with respect to the policy parameters (i.e. the policy gradient) is 
\begin{align}
    \nabla_w U(w) = \sum_{k=0}^{N-1} \EE_{x_k|\pi_w,x_0} \[ \nabla_w \mu_{k,w_k}(x_k) \nabla_{d_k} Q^{\pi_w}_k(x_k,d_k)\Big|_{d_k=\mu_{k,w_k}(x_k)} \].\label{eq:pg_theorem}
\end{align}
\end{theorem}
We provide a proof in \cref{app:pg_derive}, which follows the proof in \cite{silver2014deterministic} for a general infinite-horizon MDP. 

\subsection{Numerical Estimation of the Policy Gradient}
\label{ss:PG_numerical}

The PG \cref{eq:pg_theorem} generally cannot be evaluated in closed form, and needs to be approximated numerically. We propose a Monte Carlo (MC) estimator:
\begin{align}
    \label{eq:policy_grad}
    \nabla_w U(w) \approx \frac{1}{M} \sum_{i=1}^M \sum_{k=0}^{N-1} \nabla_w \mu_{k,w_k}(x^{(i)}_k) \nabla_{d^{(i)}_k} Q^{\pi_w}_k(x^{(i)}_k,d^{(i)}_k)\Big|_{d^{(i)}_k=\mu_{k,w_k}(x^{(i)}_k)}
\end{align}
where superscript indicates the $i$th episode (i.e. trajectory instance) generated from MC sampling. Note that the \emph{sampling} only requires a given policy and does not need any Q-function. Specifically, for the $i$th episode, we first sample a hypothetical ``true'' $\theta^{(i)}$ from the prior belief state $x_{0,b}$ and freeze it for the remainder of this episode---that is, all subsequent $y_k^{(i)}$ will be generated from this $\theta^{(i)}$.
We then compute $d_k^{(i)}$ from the current policy $\pi_w$, sample $y_k^{(i)}$ from the likelihood $p(y_k|\theta^{(i)},d_k^{(i)},I_k^{(i)})$, and repeat for all experiments $k=0,\dots,N-1$. The same procedure is then repeated for all episodes $i=1,\dots,M$. The choice of $M$ can be selected based on indicators such as MC standard error, ratio of noise level compared to gradient magnitude, or the validation expected utility from sOED policies produced under different $M$. 
While we propose to employ a fixed sample $\theta^{(i)}$ for the entire $i$th episode, one may also choose to resample $\theta_k^{(i)}$ at each stage $k$ from the updated posterior belief state $x_{k,b}^{(i)}$.
These two approaches are in fact equivalent, since from factoring out the expectations we have
\begin{align}
    \label{eq:equivalency_sample_theta}
    U(w) &= \EE_{y_0,...,y_{N-1}|\pi_w,x_0}\[\sum_{k=0}^{N-1}g_k(x_k,d_k,y_k)+g_N(x_N)\] \nonumber \\
    &= \EE_{\theta|x_{0,b}} \EE_{y_0|\pi_w,\theta,x_0} \EE_{y_1|\pi_w,\theta,x_0,y_0} \cdots \nonumber\\
    &\qquad \qquad \qquad \cdots \EE_{y_{N-1}|\pi_w,\theta,x_0,y_0,\dots,y_{N-2}} \[\sum_{k=0}^{N-1}g_k(x_k,d_k,y_k)+g_N(x_N)\] \\
    &= \EE_{\theta_0|x_{0,b}} \EE_{y_0|\pi_w,\theta_0,x_0} \EE_{\theta_1|x_{1,b}} \EE_{y_1|\pi_w,\theta_1,x_{1}} \cdots \nonumber\\
    & \qquad \qquad \qquad \cdots\EE_{\theta_{N-1}|x_{N-1,b}} \EE_{y_{N-1}|\pi_w,\theta_{N-1},x_{N-1}} \[\sum_{k=0}^{N-1}g_k(x_k,d_k,y_k)+g_N(x_N)\],
\end{align}
where the second equality corresponds to the episode-fixed $\theta^{(i)}$, and the last equality corresponds to the resampling of $\theta_k^{(i)}$. The former, however, is computationally easier, since it does not require working with the intermediate posteriors.

From \cref{eq:policy_grad}, the MC estimator for PG entails computing the gradients $\nabla_w \mu_{k,w_k}(x^{(i)}_k)$ and $\nabla_{d^{(i)}_k} Q^{\pi_w}_k(x^{(i)}_k,d^{(i)}_k)$. While the former can be obtained through the parameterization of the policy functions, the latter typically requires parameterization of the Q-functions as well. We thus parameterize both the policy and Q-functions, arriving at an actor-critic method. Furthermore, we adopt the approaches from Deep Q-Network (DQN)~\cite{mnih2015human} and 
DDPG~\cite{lillicrap2015continuous}, and use DNNs to approximate the policy and Q-functions. We present these details next.

\subsubsection{Policy Network}
\label{sec:policy_net}

Conceptually, we would need to construct individual DNNs $\mu_{k,w_k}$ to approximate $\mu_{k} : \CX_k \mapsto \CD_k$ for each $k$. Instead, we choose to combine them together 
into a single function $\mu_{w}(k, x_k)$, which then requires only a single DNN for the entire policy at the cost of a higher input dimension. Subsequently, the $\nabla_w \mu_{k,w_k}(x^{(i)}_k)=\nabla_w \mu_{w}(k,x^{(i)}_k)$ term from \cref{eq:policy_grad} can be obtained via back-propagation. Below, we discuss the architecture design of such a DNN, with particular focus on its input layer.

For the first input component, i.e. the stage index $k$,
instead of passing in the integer directly we opt to use one-hot encoding that takes the form of a unit vector:
\begin{align}
    k \qquad \longrightarrow \qquad e_{k+1}=[0,\dots,0,\underbrace{1}_{(k+1)\rm{th}},0,\dots,0]^T.
\end{align}
We choose one-hot encoding because the stage index
is an ordered categorical variable instead of a quantitative variable (i.e. it has notion of ordering but no notion of metric). Furthermore, these unit vectors are always orthogonal, which we observed to offer good overall numerical performance of the policy network. The tradeoff is that the dimension of representing $k$ is increased from 1 to $N$.

For the second component, i.e. the state $x_k$ (including both $x_{k,b}$ and $x_{k,p}$), we represent it in a nonparametric manner as discussed in \cref{sec:math_formulation}:
\begin{align}
x_k \qquad \longrightarrow \qquad I_k=\{d_0,y_0,\dots,d_{k-1},y_{k-1}\}.
\end{align}
To accommodate states up to stage $(N-1)$ (i.e. $x_{N-1}$), we use a fixed total dimension of $(N-1)(N_d+N_y)$ for this representation, where for $k < (N-1)$ the entries for $\{d_l, y_l \,|\, l \geq k\}$ (experiments that have not happened yet) are padded with zeros (see \cref{eq:NN_input}). 
In addition to providing a 
state representation without any approximation, another major advantage of such nonparametric form can be seen under the terminal formulation in \cref{eq:terminal_info_gN}, where now none of the intermediate belief states (i.e. $x_{k,b}$ for $k<N$) needs to be computed since the policy network can directly take input of $I_k$. 
As a result, only a single final Bayesian inference conditioned on all experiments and all observations needs be performed at the end of each episode. The number of Bayesian inference calculations is greatly reduced.

Putting together the two input components, the overall input layer for the policy network $\mu_{w}(k,x_k)$, when evaluating at $(k,x_k)$, has the form
\begin{align}
    \label{eq:NN_input}
    [ \underbrace{e_{k+1}}_{N},\overbrace{d_0}^{N_d},\dots,d_{k-1},\underbrace{0,\dots,0}_{N_d(N-1-k)},\overbrace{y_0}^{N_y},\dots,y_{k-1},\underbrace{0,\dots,0}_{N_y(N-1-k)} ] ^ T,
\end{align}
where we also indicate the zero-paddings for the entries corresponding to future experiments $l \geq k$. 
The overall input layer has a total dimension of $N + (N-1)(N_d+N_y)$, which is linear in $N$, $N_d$, and $N_y$. 
The remainder of the policy network is relatively straightforward. The output layer is an $N_d$-dimensional vector representing $d_k$, and the network architecture can be chosen by the user. 
We have experimented with dense layers, and experience suggests 2-3 hidden layers often achieve good performance for our numerical cases. More systematic hyperparameter tuning for DNNs can also be employed to optimize the architecture, but not pursued in this paper. 

We end the introduction of the policy network by emphasizing that $\mu_{w}(k,x_k)$ is not trained in a supervised learning manner from training data; instead, it is updated iteratively via PG en route to maximizing $U(w)$.

\subsubsection{Q-Network}
\label{sec:Q_net}

While seeking $Q^{\pi_w}_{k,\eta_k}$ (parameterized by $\eta_k$) that approximates $Q^{\pi_w}_k : \CX_k \times \CD_k \mapsto \RR$ for $k=0,\dots,N-1$,
we also combine them into a single function $Q^{\pi_w}_{\eta}(k,x_k,d_k)$ in a similar manner as the policy network; we call $Q^{\pi_w}_{\eta}$ the Q-network.
Subsequently, the $\nabla_{d^{(i)}_k} Q^{\pi_w}_k(x^{(i)}_k,d^{(i)}_k)$
term from \cref{eq:policy_grad} can be approximated by 
$\nabla_{d^{(i)}_k} Q^{\pi_w}_{\eta}(k,x^{(i)}_k,d^{(i)}_k)$, which can now also be obtained via back-propagation. 
The input layer then takes the same form as the policy network, except we augment extra entries for $d_k$ as well. The overall input dimension is then $N+(N-1)(N_d+N_y)+N_d$. The network output is a scalar.

The Q-network is trained in a supervised learning manner from the MC episodes generated for \cref{eq:policy_grad}, by finding $\eta$ that minimizes the following loss function built based on
\cref{eq:action_bellman2}:
\begin{align}
    \label{eq:value_loss}
    \CL(\eta) = \frac{1}{M} \sum_{i=1}^M \sum_{k=0}^{N-1} \[ Q^{\pi_w}_{\eta}(k,x^{(i)}_k,d^{(i)}_k) - \(g_k(x^{(i)}_k,d^{(i)}_k,y^{(i)}_k) + Q^{\pi_w}_{k+1}(x^{(i)}_{k+1},d^{(i)}_{k+1})\) \]^2
\end{align}
where $d^{(i)}_k = \mu_{w}(k,x^{(i)}_k)$ 
and $Q^{\pi_w}_N(x^{(i)}_N,\cdot) = g_N(x^{(i)}_N)$. It is worth noting that $Q^{\pi_w}_{k+1}(x^{(i)}_{k+1},d^{(i)}_{k+1})$ 
does not depend on $\eta$, but in practice is often approximated by  $Q^{\pi_w}_{\eta}(k+1,x^{(i)}_{k+1},d^{(i)}_{k+1})$ ({for $k=0,\dots,N-2$})\footnote{The use of an approximate Q-value in the next (i.e. $k+1$) stage rather than expanding further with $g_{k+1}$, $g_{k+2}$, etc. renders this a \emph{one-step lookahead} approach. This is not to be confused with greedy or myopic design, which does not include any future value term.}. When minimizing the loss, the gradient contribution with respect to $\eta$ from this term is therefore ignored.

\subsubsection{Evaluation of Kullback-Leibler Rewards}
\label{sec:numerical_belief_state}

A final step needed to construct the Q-network following \cref{eq:value_loss} (and in turn, the policy network) is the ability to evaluate our immediate and terminal rewards $g_k$ and $g_N$. 
Having established the equivalence of terminal and incremental formulations in \cref{sec:information_gain}, we focus on the former since it requires fewer KL divergence calculations with only the KL in $g_N$ needed at the end of each episode. With the nonparametric state representation using $I_k$ (\cref{sec:policy_net}), we do not need to explicitly update the Bayesian posteriors throughout the intermediate experiments. Instead, we only need a single Bayesian inference to obtain $p(\theta|I_N)$, and use it to estimate the KL divergence \cref{eq:terminal_info_gN}. 

In general, the posteriors will be of non-standard distributions and the KL divergence 
must be approximated numerically. 
For the small $N_{\theta}$ (e.g., $\leq 4$) examples presented in this paper, we use a grid discretization of the $\theta$-space and estimate its posterior PDF pointwise; in this work, we always use a uniform grid with 50 nodes in each dimension for evaluation, and 20 nodes for training only in the higher-dimensional $N_{\theta}=4$ case to speed up the computation (spot-checking between 20 and 50 nodes indicated their results to be close). 
However, the exponential growth of grid points with $N_{\theta}$ would require higher dimensional problems to seek alternative methods, such as Markov chain Monte Carlo (MCMC) with kernel density estimation or likelihood-free ratio estimation \cite{thomas2021likelihood}, variational inference \cite{blei2017variational} and transport maps \cite{Huan2015}. These will be important future directions of our research.

\subsubsection{Exploration Versus Exploitation}

The tradeoff between exploration and exploitation is an important consideration in RL, especially in an uncertain environment. 
Exploration searches unexplored or under-explored regions that may contain good policies (i.e. global search), and invests for long-term performance. Exploitation focuses on region deemed promising based on our current knowledge (i.e. local search), thus targets short-term performance. Insufficient exploration may strand the policy search in a local optimum, and insufficient exploitation may lack 
convergence. A mixed strategy to balance 
exploration and exploitation is prudent \cite{burnetas1997optimal,li2017deep}, such as through the commonly used epsilon-greedy technique \cite{sutton2018reinforcement} and many other advanced methods.

In this work, we inject exploration by adding a perturbation to our deterministic policy. 
We employ this exploration \emph{only} when generating the MC episodes in \cref{eq:policy_grad} for estimating the PG during training, and nowhere else (including testing). Thus we view this exploration as an aid solely to the training data generation, and our policy remains deterministic. 
When this exploration perturbation is used, the design becomes:
\begin{align}
    d_k = \mu_k(x_k) + \epsilon_{\rm{explore}},
\end{align}
where $\epsilon_{\rm{explore}}\sim\CN(0,\mathbb{I}_{N_d}\sigma_{\rm{explore}}^2)$.
The perturbed $d_k$ should also be truncated by any design constraints to remain within $\CD_k$. The value of $\sigma_{\rm{explore}}$ reflects the degree of exploration versus exploitation, and should be selected based on the problem context. For example, a reasonable approach is to set a large $\sigma_{\rm{explore}}$ early in the algorithm and reduce it gradually. More advanced techniques have been proposed to reach a better exploration, for instance, by adding noise to the policy network parameters instead of the design variable \cite{plappert2017parameter,fortunato2017noisy}; however, these strategies are beyond the scope of our paper.

\subsection{Simplification to Batch and Greedy Designs}
\label{sec:PG-based_batch_greedy}

In \cref{sec:subopt_design}, we illustrated batch and greedy designs to be suboptimal cases of the sOED. Similarly, we can reduce the PG-sOED formulation, with very minor modifications, to arrive at PG-based batch and greedy designs. 

For batch design, we simply replace the input layer of the policy network to contain only the one-hot encoding terms $e_{k+1}$. The resulting policy structure thus only maps from the stage index $k$ to a design, and $d_k$ does not depend on the state. 
For greedy design, we 
use the incremental formulation
and retain only the immediate reward term. The Q-function then becomes $Q^{\pi}_k(x_k,d_k)=\EE_{y_k|x_k,d_k}[g_k(x_k,d_k,y_k)]$, and the loss in \cref{eq:value_loss} simplifies to
\begin{align*}
        {\CL^{\mathrm{gr}}}(\eta) = \frac{1}{M} \sum_{i=1}^M \sum_{k=0}^{N-1} \[ Q^{\pi_w}_{\eta}(k,x^{(i)}_k,d^{(i)}_k) - g_k(x^{(i)}_k,d^{(i)}_k,y^{(i)}_k) \]^2.
\end{align*}

\subsection{Pseudocode for the Overall Algorithm}

We present the detailed algorithm for PG-sOED in \cref{alg:PG-sOED}. 
We re-emphasize that the exploration perturbation is only used in generating the MC episodes on line 5, but not used anywhere else including when evaluating the policy after it is constructed. 
Furthermore, with our DNN parameterization of the policy and Q-functions (\cref{sec:policy_net,sec:Q_net}), $w$ and $\eta$ are their respective DNN parameter weights. Hence, DNNs computations are encountered in \cref{alg:PG-sOED} wherever $w$ and $\eta$ appear.

\begin{algorithm}
\caption{The PG-sOED algorithm.}
\label{alg:PG-sOED}
\begin{algorithmic}[1]
\STATE{Define all components in \cref{sec:math_formulation};}
\STATE{Set initial state $x_0$, policy updates $L$, MC sample size $M$, policy and Q-network architectures, learning rate $\alpha$ for policy update, exploration scale $\sigma_{\rm{explore}}$;
}
\STATE{Initialize policy and Q-network parameters $w$ and $\eta$;}
\FOR{$l=1,\dots,L$}
\STATE{Simulate $M$ episodes: sample $\theta\sim x_{0,b}$, and then for $k=0,\dots,N-1$ sample $d_k=\mu_{w}(k,x_k)+\epsilon_{\rm{explore}}$} and $y_k\sim p(y_k|\theta,d_k,I_k)$;
\STATE{Store the full information vectors from all episodes $\{I_N^{(i)}\}_{i=1}^M$}, from which the intermediate $\{I_1^{(i)}, I_2^{(i)},\ldots,I_{N-1}^{(i)}\}$ can also be formed trivially;
\STATE{Compute and store immediate and terminal rewards for all episodes $\{g_k^{(i)}\}_{i=1}^M$, $k=0,\dots,N$;}
\STATE{Update $\eta$ by minimizing the loss in \cref{eq:value_loss};}
\STATE{Update $w$ by gradient ascent: $w = w + \alpha \nabla_wU(w)$, where $\nabla_wU(w)$ is estimated through \cref{eq:policy_grad};}
\STATE{(Optional) Reduce $\alpha$ and $\sigma_{\rm{explore}}$;}
\ENDFOR
\STATE{Return optimized policy $\pi_{w}$;}
\end{algorithmic}
\end{algorithm}

\section{Numerical Results}
\label{sec:results}

We present two examples to illustrate different aspects of PG-sOED. The first is a linear-Gaussian problem (\cref{sec:ex_linGau}) that offers a closed form solution due to its conjugate prior form. This problem serves as a benchmark, where we validate the optimal policy and expected utility obtained by PG-sOED against the analytic solution. We also illustrate the superior computational speed of PG-sOED over an existing ADP-sOED baseline.
The second example entails a problem of contaminant source inversion in a convection-diffusion field (\cref{sec:ex_sourceInv}). It further divides into three cases: case 1 compares PG-sOED to  greedy design, and cases 2 and 3 compare PG-sOED to both greedy and batch designs. This example thus demonstrates the advantages of PG-sOED over greedy and batch designs, and its ability to accommodate expensive forward models with nonlinear physics and dynamics. 

\subsection{Linear-Gaussian Benchmark}
\label{sec:ex_linGau}
\subsubsection{Problem Setup}
We adopt the linear-Gaussian problem from~\cite{Huan2016} as a benchmark case for validating PG-sOED. The forward model is linear in $\theta$, and corrupted with an additive Gaussian observation noise $\epsilon_k \sim \CN(0,1^2)$:
\begin{align}
    y_k = G(\theta,d_k) + \epsilon_k = \theta d_k + \epsilon_k.
\end{align}
We design $N=2$ experiments, with prior $\theta\sim\CN(0,3^2)$, and design constrained in $d_k \in [0.1,3]$. The resulting conjugate form renders all subsequent posteriors to be analytically Gaussian, thus allowing us to compute the optimal policy in closed form. 
There is no  physical state for this problem. 
The stage rewards and terminal reward are chosen to be
\begin{align}
    g_k(x_k,d_k,y_k) &= 0, \quad k=0,1 \\
    g_N(x_N) &= \DKL\( p(\cdot|I_N)\,||\,p(\cdot|I_0) \) - 2\( \ln{\sigma_N^2} - \ln{2} \)^2
\end{align}
where $\sigma_N^2$ represents the variance of the final belief state, and the additive penalty in the terminal reward is purposefully inserted 
to make the design problem more challenging.
We solve this sOED problem both by ADP-sOED \cite{Huan2016} and PG-sOED. ADP-sOED adopts the same setting in \cite{Huan2016}. For PG-sOED, we set 
$L=100$, 
$M=1000$, 
$\alpha=0.15$, 
and $\sigma_{\rm{explore}}=0.2$ {that also decreases by
a factor of $0.95$ per policy update. Both the policy network and Q-network contain two hidden layers with ReLU activation, and each hidden layer has 80 nodes.} In particular, we have observed even a low $M=10$ yielded similar performance, but $M=1000$ is used here to further reduce the effect of MC error in the demonstration. 

\subsubsection{Results}

Due to the conjugate form, we can obtain all posteriors in closed form, and find the (non-unique) optimal policies analytically~\cite{Huan2015,Huan2016}. To evaluate the policies found by ADP-sOED and PG-sOED, we sample $10^4$ episodes using their final policies and compute their total rewards. 
ADP-sOED yields a mean total reward of 
$0.775\pm0.006$ and PG-sOED also 
$0.775\pm0.006$, where the $\pm$ is the MC standard error. Both match extremely well with the analytic result $U(\pi^\ast)\approx0.783$, where the discrepancy (due to MC sampling {and grid discretization of the posterior}) is within two standard errors. These results thus support that both ADP-sOED and PG-sOED have found the optimal policy.

\begin{figure}[htbp]
  \centering
  \subfloat[Reward history]{\label{fig:LinGau_Cnvg_reward}\includegraphics[width=0.49\linewidth]{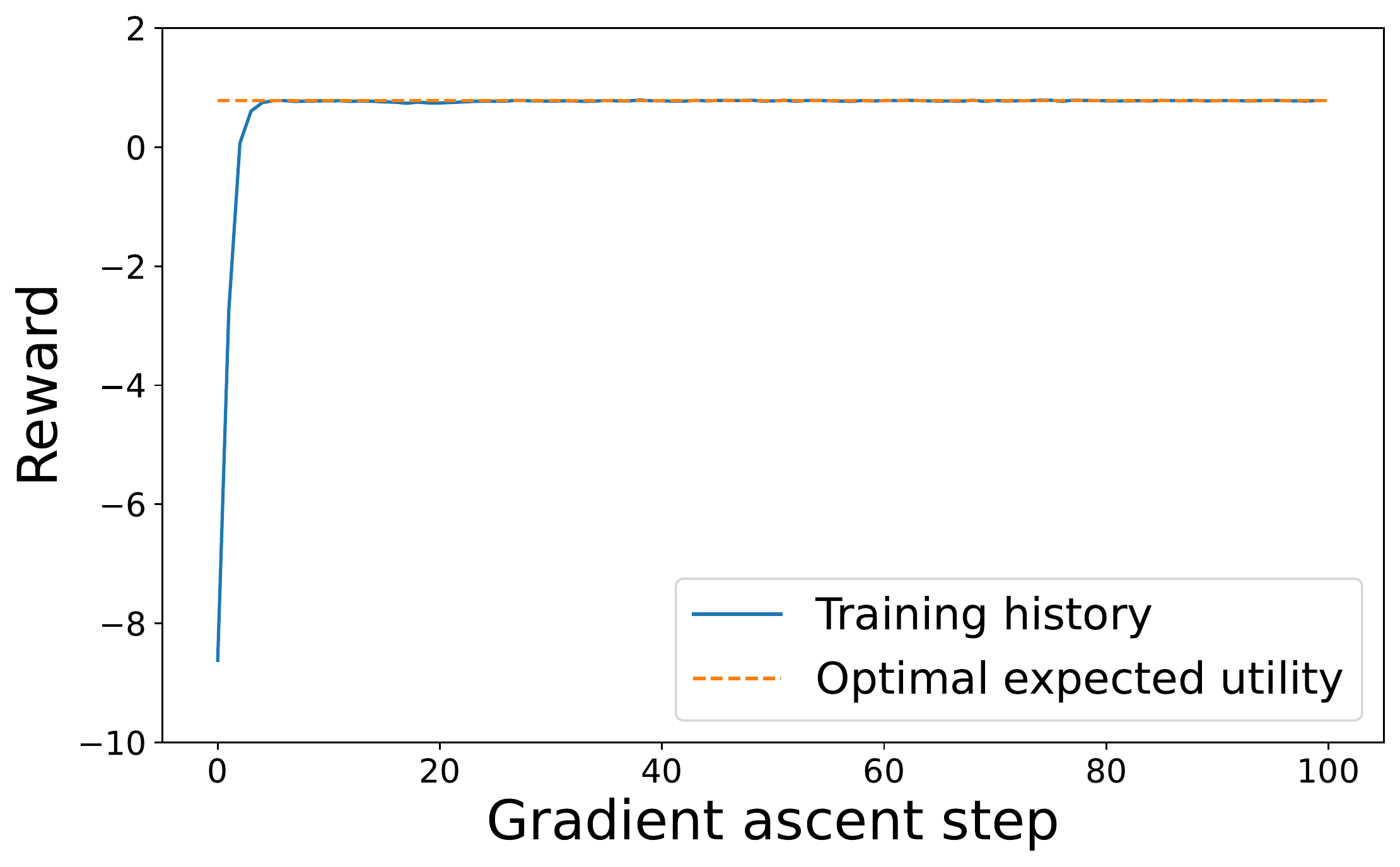}}
  \subfloat[Residual history $|U(\pi^\ast)-U(\pi)|$]{\label{fig:LinGau_Cnvg_residual}\includegraphics[width=0.49\linewidth]{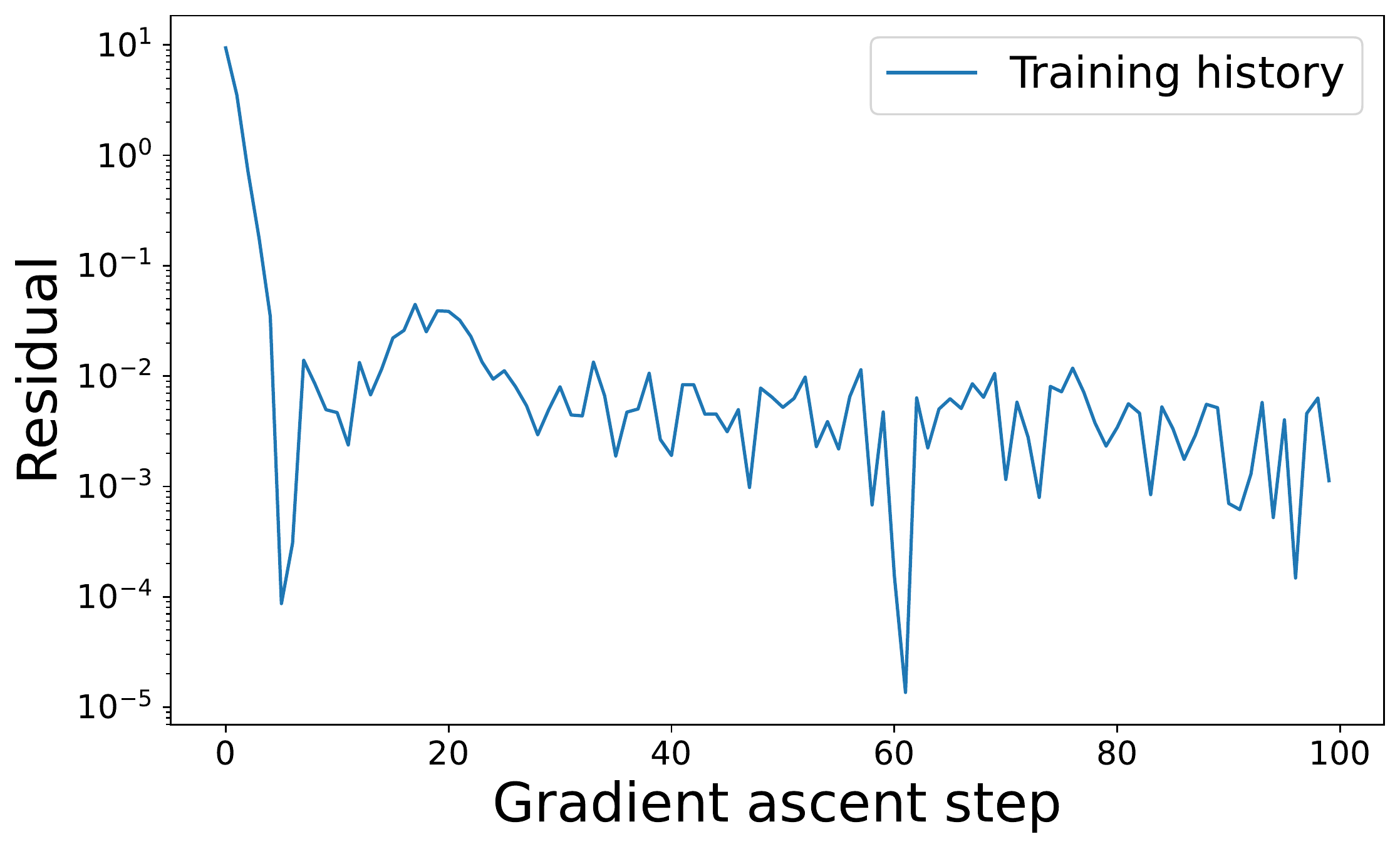}}
  \caption{Convergence history of PG-sOED.}
  \label{fig:LinGau_Cnvg}
  
\end{figure}

\Cref{fig:LinGau_Cnvg_reward,fig:LinGau_Cnvg_residual} present the convergence history for the expected utility and residual ($|U(\pi^\ast)-U(w)|$) as a function of the PG-sOED iterations. 
The convergence is rapid, reaching over 3 orders of magnitude reduction of the residual 
within 30 iterations. 
The much lower initial expected utility (around $-8.5$) also indicates that a random policy (from random initialization) performs much worse than the optimized policy.

\cref{tab:ADPvsPG} compares the computational costs between ADP-sOED and PG-sOED for this linear-Gaussian problem, obtained using a single 2.6 GHz CPU on a MacBook Pro laptop.
The timing figures are from 30 gradient ascent updates for PG-sOED in the training stage, and 1 policy update (minimum needed) for ADP-sOED. 
PG-sOED produces orders-of-magnitude speedups compared to ADP-sOED. Noteworthy is the extremely low testing time (i.e. using the policy online during the experimental campaign after the policy has been constructed offline), achieving $0.0002$ seconds per experiment (4 seconds per $10^4$ episodes with $N=2$ experiments per episode). 
The advantage is due to ADP-sOED being a value-based approach where each policy evaluation needs to solve a (stochastic) optimization problem which in turns requires many forward model evaluations; in contrast, PG-sOED only requires a single forward pass of its policy-network free of any forward model runs. The online speed make the PG-sOED an excellent candidate for situations where fast, possibly real-time design responses are needed.

\begin{table}[hbt]
    \centering
    \caption{Comparison of computational costs between ADP-sOED and PG-sOED.
        }
    \begin{tabular}{|c| c c |c|}
    \hline
        & \textbf{Training time (s)} & \textbf{Forward model evaluations} & \textbf{Testing time (s)} \\
    \hline
    ADP-sOED & 837     & $5.3 \times 10^8$     & 24,396     \\
    PG-sOED & 24     & $3.1 \times 10^6$     & 4     \\
    \hline
    \end{tabular}
    \label{tab:ADPvsPG}
\end{table}

\subsection{Contaminant Source Inversion in Convection-Diffusion Field}
\label{sec:ex_sourceInv}

\subsubsection{Problem Setup}

\begin{figure}[htbp]
  \centering
  \includegraphics[width=1.0\linewidth]{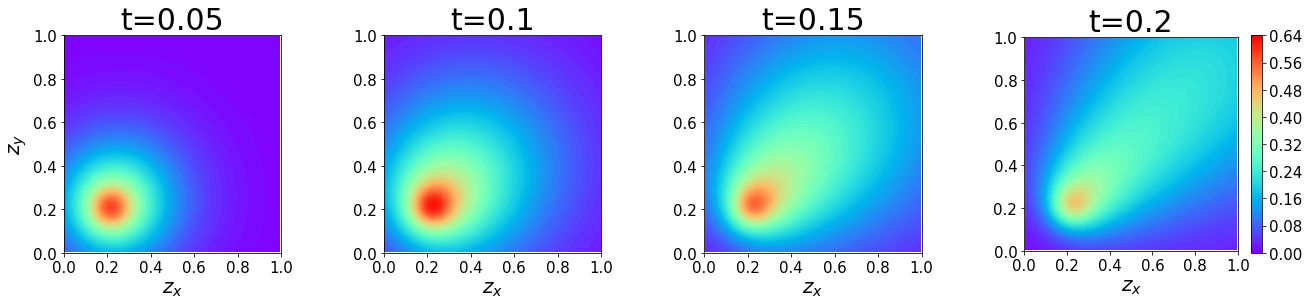}
    \caption{Sample numerical solution of the concentration field $G$ at different time snapshots. The solution is solved in a wider computational domain $[-1,2]^2$ but displayed here in a region of interest $[0,1]^2$. In this case, $\theta=[0.210,0.203,0.05,2]$ and the convection grows over time with $u_x=u_y=10t/0.2$. Hence, isotropic diffusion dominates early on and the plume stretches towards the convection direction with time. 
    }
  \label{fig:Source_exp}
\end{figure}

The next example entails mobile sensor design in a convection-diffusion field (e.g., of a chemical contaminant plume) to take concentration measurements in order to infer the plume source location. We model the contaminant field in a two-dimensional square domain, where the contaminant concentration $G$ at spatial location $z=[z_x,z_y]$ and time $t$ is constrained by the convection-diffusion PDE:
\begin{align}
    \frac{\partial G(z,t;\theta)}{\partial t} = \nabla^2 G-{u}(t) \cdot \nabla G + S(z,t;\theta), \qquad z\in [z_L,z_R]^2, \quad t>0,
    \label{e:PDE}
\end{align}
where ${u}=[u_x,u_y] \in \RR^2$
is a time-dependent convection velocity, and $\theta =[\theta_x,\theta_y, \theta_h, \theta_s] \in \RR^4$ 
{is the source parameter} residing within the source function
\begin{align}
    S(z,t;\theta) = \frac{\theta_s}{2\pi \theta_h^2} \exp\( -\frac{(\theta_x-z_x)^2 + (\theta_y-z_y)^2}{2 \theta_h^2} \)
\end{align}
with $\theta_x$ and $\theta_y$ denoting the source location, and $\theta_h$ and $\theta_s$ denoting the source width and source strength. 
The initial condition is $G(z,0;\theta)=0$, and homogeneous Neumann boundary condition 
is imposed for all sides of the square domain. 
We solve the PDE numerically using second-order finite volume method on a uniform grid of size $\Delta z_x = \Delta z_y = 0.01$ and {a second-order fractional step method for} time-marching with stepsize $\Delta t = 5.0 \times 10^{-4}$. 
For example, \cref{fig:Source_exp} illustrates the solution $G$ for such a convection-diffusion scenario where the convection speed increases over time. 

For the design problem, we have a vehicle with sensing equipment for measuring the contaminant concentration $G$ that can be relocated at fixed time intervals. We seek to determine where we should relocate this vehicle such that its measurements can lead to the best inference of the source parameter $\theta$.
We consider $N$ measurement opportunities respectively at time $t_k$ for $k=0,\dots,N-1$. The vehicle starts with initial belief state $x_{0,b}=(\theta|I_0)$ (i.e. prior on $\theta$) and initial physical state $x_{0,p}$ (i.e. vehicle location). The design variable is the displacement of the vehicle from the current location and constrained $d_k \in [d_L, d_R]^2$ to reflect the range of vehicle movement. The physical state is then updated via
\begin{align}
    x_{k+1,p} = x_{k,p} + d_k.
\end{align}
At the new physical location, a noisy measurement of the contaminant concentration is obtained in the form
\begin{align}
    y_k = G(z=x_{k+1,p},t_k;\theta) + \epsilon_k \( 1+\vert G(x_{k+1,p},t_k;\theta) \vert \)
\end{align}
where 
$\epsilon_k \sim \CN(0,\sigma_\epsilon^2)$, thus the observation noise is affected by the signal magnitude. 
Once the new measurement is acquired, the belief state is updated from $x_{k,b}=(\theta|I_k)$ to $x_{k+1,b}=(\theta|I_{k+1})$ through Bayes' rule. The reward functions are 
\begin{align}
    g_k(x_k,d_k,y_k) &= - c_q f_c(d_k), \quad k=0,\dots,N-1 \\
    g_N(x_N) &= 
                        \DKL\( p(\cdot|I_N)\,||\,p(\cdot|I_0) \).
\end{align} 
In particular, the immediate reward reflects a cost on the vehicle movement considering the effect of the wind, {where $f_c(d_k)$ denotes the cost function on sensor movement.}

We further set up 3 cases for this source inversion problem with their settings summarized in \cref{tab:Source_setup}. 
Case 1 is diffusion-only and cases 2 and 3 are convection-diffusion. {In cases 1 and 2, $N=2$ experiments will be designed, and the source width $\theta_h$ and source strength $\theta_s$ are known. In case 3, $N=4$ experiments will be designed, and $\theta_h$ and $\theta_s$ are treated as unknown parameters}.
For case 1, contaminant source is off ($s=0$) initially and activated ($s=2$) at $t=0.16$. The first experiment ($t_0=0.15$) thus takes place when there is no contaminant and only the second ($t_1=0.32$) encounters the plume. We anticipate no immediate gain from the first experiment but it may be used to set up a better second experiment via lookahead. We use case 1 to highlight the difference between sOED and greedy design. 
Case 2, by including time-dependent convection, emphasizes the value of feedback adaptation to environmental change and dynamics. We use case 2 to compare sOED with both greedy and batch designs. 
{Case 3 features a longer horizon and a higher dimensional parameter space, and with a penalty on the sensor movement. We use case 3 to demonstrate the usage of sOED in more complicated problems.}
All batch and greedy designs are implemented following \cref{sec:PG-based_batch_greedy}. 
For PG-sOED, we set $L=300$, $M=1000$, $\alpha=0.01$ {with the Adam optimizer \cite{kingma2014adam}}, and $\sigma_{\rm{explore}} = 0.05$. From experimentation with different $M$'s, we have found $M=1000$ to be suitable here while lower settings such as $M=100$ starts to {produce noisier results}.
Finally, $10^4$ episodes are simulated for evaluating the performance of different design policies.

\begin{table}[hbt]
    \centering
    \caption{Setup for the three cases in the contaminant source inversion problem.}
    \begin{tabular}{|c|c|c|c|}
    \hline
    & \textbf{Case 1} & \textbf{Case 2} & \textbf{Case 3}\\
    \hline
    Number of experiments & \multicolumn{2}{c|}{$N=2$} & $N=4$ \\
    \cline{2-4}
    Prior of $\theta_x$ and $\theta_y$ & \multicolumn{3}{c|}{$\theta_x,\theta_y \sim \CU([0,1])$} \\
    \cline{2-4}
    Prior of $\theta_h$ & \multicolumn{2}{c|}{$\theta_h=0.05$} & $\theta_h \sim \CU([0.02, 0.1])$ \\
    \cline{2-4}
    Prior of $\theta_s$ & $\theta_s= \left\{ \begin{array}{c}
     0\,\, \mathrm{if} \,\, t<0.16 \\ 2\,\, \mathrm{if} \,\, t\geq 0.16 \end{array}
     \right.$ & $\theta_s=2$ & $\theta_s \sim \CU([0, 5])$ \\
    \cline{2-4}
    Initial physical state & \multicolumn{3}{c|}{$x_{0,p} = [0.5,0.5]$} \\
    \cline{2-4}
    Design constraint & \multicolumn{2}{c|}{$d_k \in [-0.25,0.25]^2$} & $x_{k,p} \in [0,1]^2$ \\
    \cline{2-4}
    Computational domain  & $z_L=0$, $z_R=1$ & \multicolumn{2}{c|}{$z_L=-1$, $z_R=2$} \\
    \cline{2-4}
    Experiment time & $t_0=0.15$, $t_1=0.32$ & $t_0=0.05$, $t_1=0.2$ & $t_k=0.05(k+1)$ \\
    \cline{2-4}
    Velocity field & $u_x=u_y=0$ & \multicolumn{2}{c|}{$u_x=u_y=10t/0.2$} \\
    Noise scale & $\sigma_\epsilon = 0.1$ & \multicolumn{2}{c|}{$\sigma_\epsilon = 0.05$} \\\cline{2-4}
    Cost function $f_c(d_k)$ & \multicolumn{2}{c|}{$\Vert d_k \Vert ^2$} & $\Vert d_k \Vert - \frac{\sqrt{2}}{40} d_k \cdot u(t_k)$ \\
    \cline{2-4}
    Cost coefficient & $c_q=0.5$ & $c_q=0$ & $c_q=0.2$ \\
    \hline
    \end{tabular}
    \label{tab:Source_setup}
\end{table}

\subsubsection{Surrogate Model}

\begin{figure}[htb]
  \centering
  \subfloat[$t=0.05$]{\label{fig:Source_surrogate_1}\includegraphics[width=0.48\linewidth]{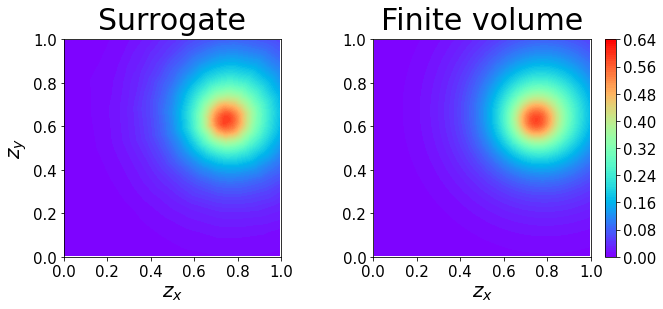}}\hspace{1em}
  \subfloat[$t=0.2$]{\label{fig:Source_surrogate_2}\includegraphics[width=0.48\linewidth]{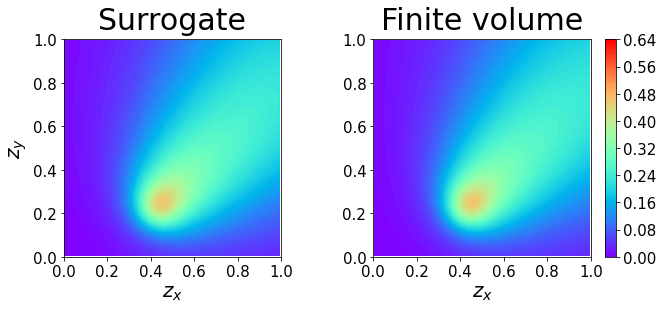}}
  \caption{Sample comparison of the concentration field $G$ at $t=0.05$ and $t=0.2$ of case 2 using the DNN surrogates (left column) and finite volume (right column). They appear nearly identical. 
    }
  \label{fig:Source_surrogate}
\end{figure}

Solving the forward model \cref{e:PDE} using finite volume is still computationally viable for PG-sOED, but expensive. 
One strategy to accelerate the computation is to employ surrogate models to replace the original forward model. 
We use DNNs to construct surrogate models of for $G(z,t_k;\theta)$ for $k=0,\dots,N-1$. 
We use the following architecture for each DNN: a 4-dimensional input layer taking $z$ and $\theta$; five hidden layers with 40, 80, 40, 20, and 10 nodes; and a scalar output $G$. 
A dataset is generated by solving for $G$ on $2000$ samples of $\theta$ drawn from its prior distribution.
These concentration values are then first restricted to only the domain that is reacheable by the vehicle (due to the design constraint), then shuffled across $\theta$ and split 80\% for training and 20\% for testing. We achieve low test mean-squared-errors of around 
$10^{-6}$
for surrogate models $G(z,t_k;\theta)$ for all three cases.
\Cref{fig:Source_surrogate} provides an example comparing the concentration contours from $t=0.05$ and $t=0.2$ of case 2 using the DNN surrogates (left column) and finite volume (right column). They appear nearly identical. 
More importantly, the surrogate models provide a significant  speedup over the finite volume solver by a factor of $10^5$. 

\subsubsection{Case 1}

\begin{figure}[htbp]
  \centering
  \includegraphics[width=0.7\linewidth]{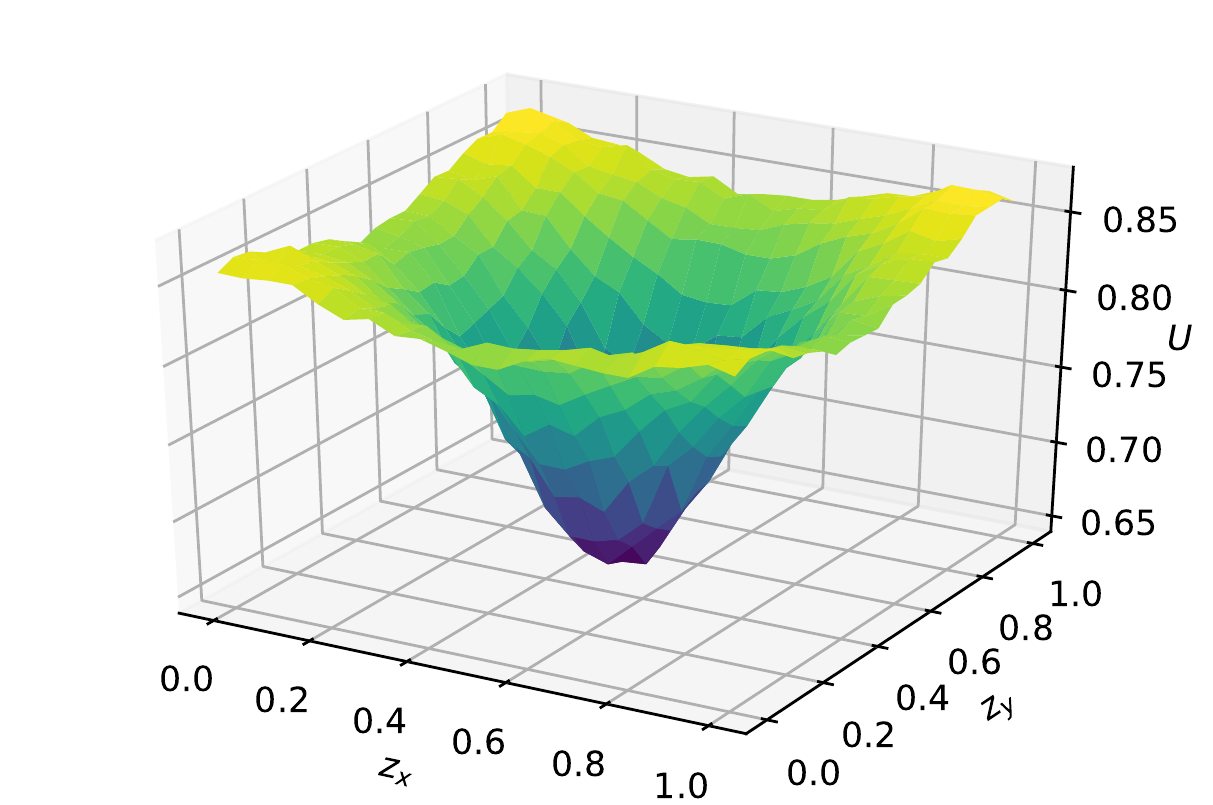}
  \caption{Case 1. Expected utility for one-experiment design at $t=0.32$. The best design locations are at the corners.}
  \label{fig:Diffusion_bestloc_util}
\end{figure}

\begin{figure}[htbp]
  \centering
  \includegraphics[width=1\linewidth]{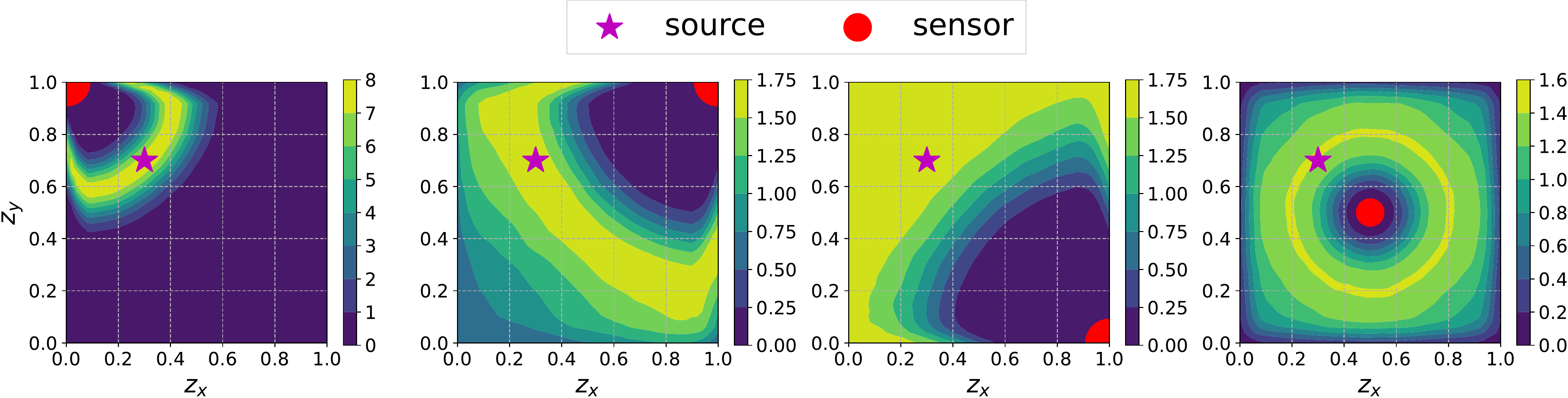}
  \caption{Case 1. Posterior PDF contours for the one-experiment design under different design locations (red dot) and a sample source location (purple star). The posteriors exhibit shapes resemble an arc of a circle, due to the isotropic nature of diffusion and the domain geometry. 
    }
  \label{fig:Diffusion_1sensor_post}
\end{figure}

Case 1 is diffusion-only. 
Before presenting the sOED results, we first offer some intuition about high-value designs from a simpler one-experiment design. 
\Cref{fig:Diffusion_bestloc_util} illustrates the expected utility surface versus sensor location for 
a \emph{single} experiment measuring concentration at
$t=0.32$. The key insight is that high-value experiments are at the corners of the domain. 
This can be explained by the isotropic nature of diffusion process that carries information about distance but not direction, thereby leading to posterior probabilities concentrating around regions that resemble an arc of a circle (\cref{fig:Diffusion_1sensor_post}). Combined with the square domain geometry and Neumann boundary conditions, the ``covered area'' of high-probability posterior is smallest (i.e. least uncertain), averaged over all possible $\theta$ source locations, when the measurements are made at the corners of the domain.

\begin{figure}[htbp]
  \centering
  \subfloat[PG-sOED, $g_0=-0.040,\sum_{k=0}^2 g_k=2.941 $]{\label{fig:Diffusion_vsgreedy_soed1}\includegraphics[width=0.48\linewidth]{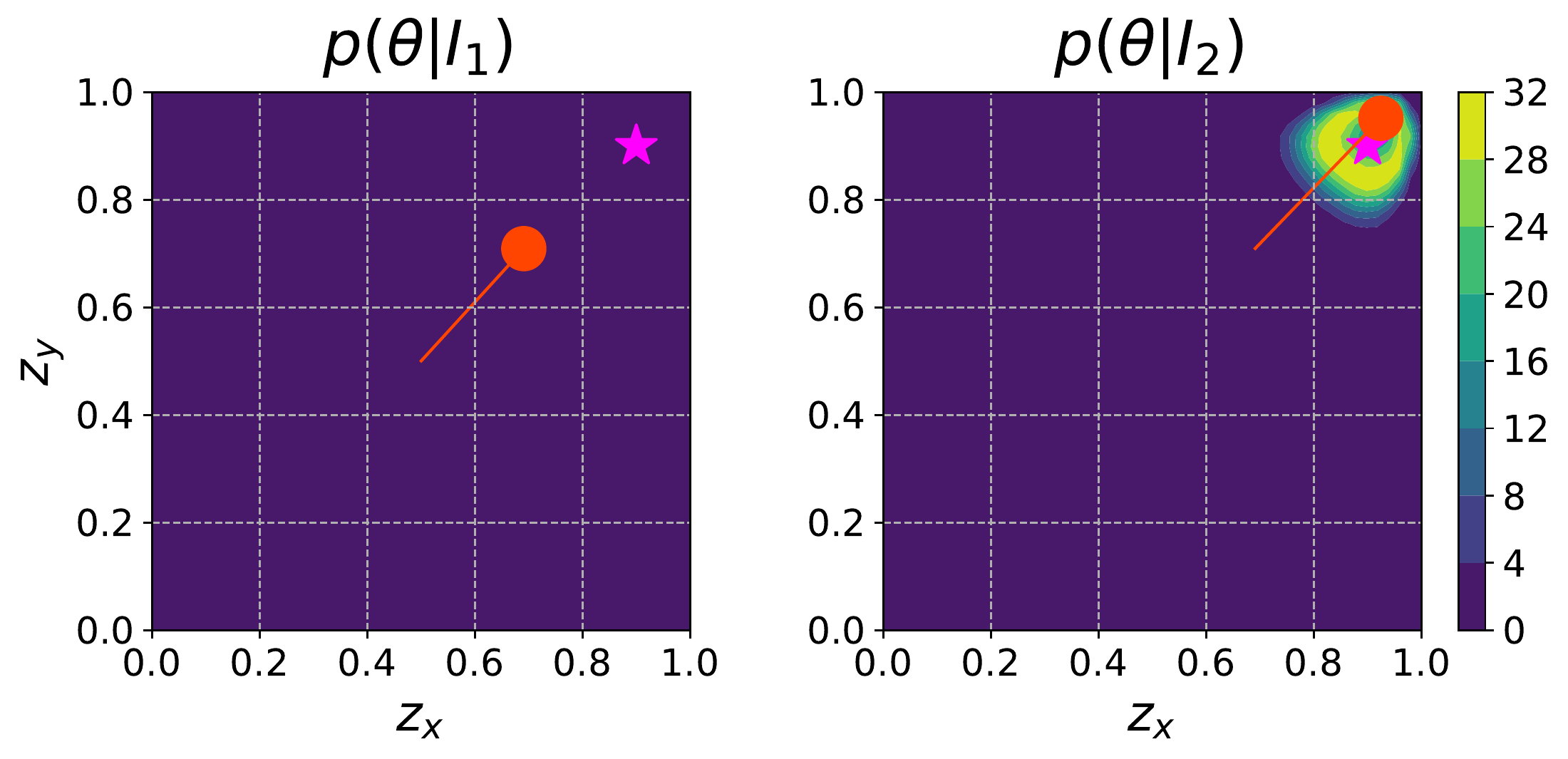}}\hspace{1em}
  \subfloat[Greedy, $g_0=-0.000, \sum_
  {k=0}^2 g_k=2.022$]{\label{fig:Diffusion_vsgreedy_greedy1}\includegraphics[width=0.48\linewidth]{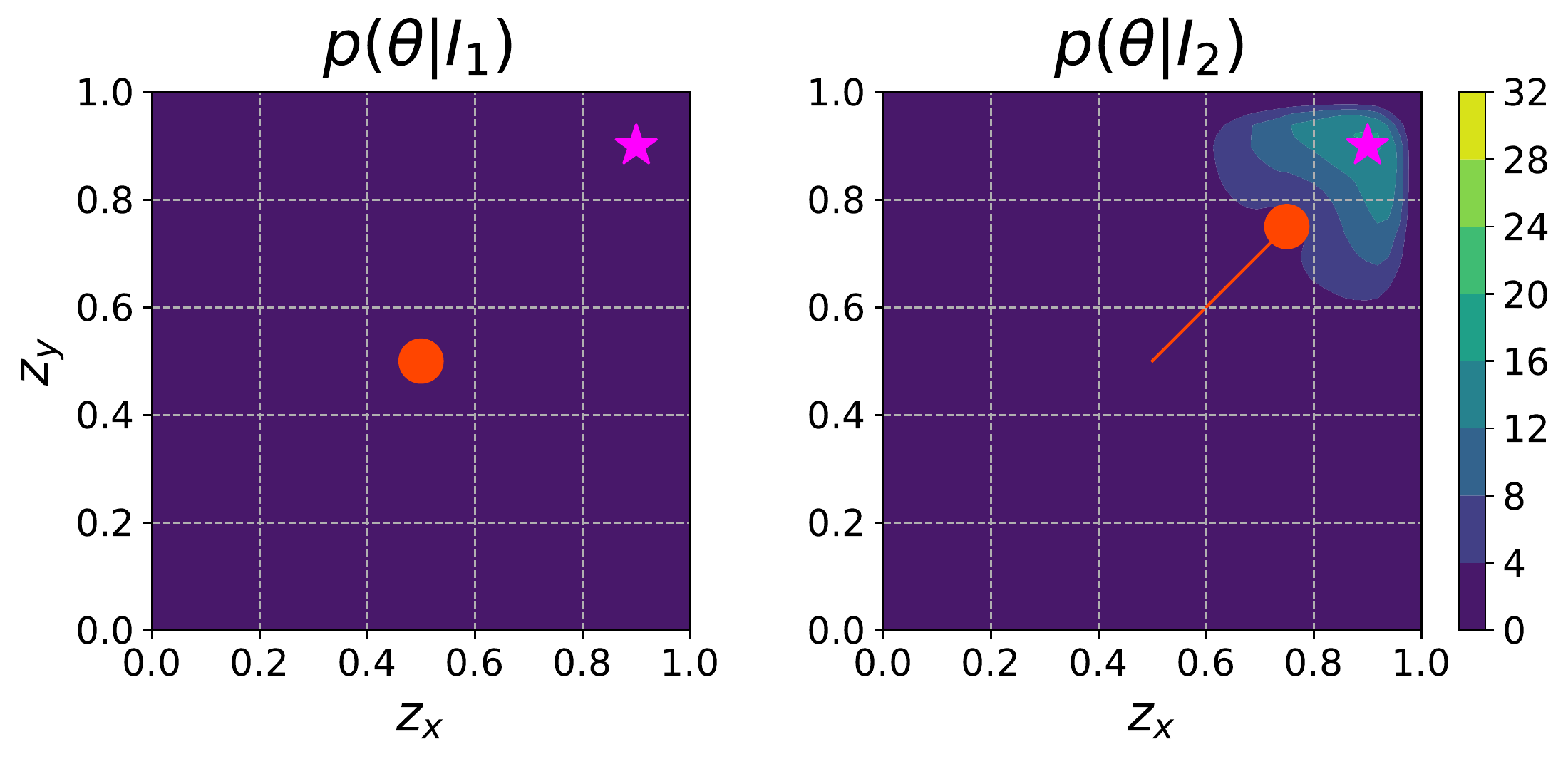}}
  \caption{Case 1. An episode instance obtained by PG-sOED and greedy design. The purple star represents the true $\theta$, red dot represents the physical state (vehicle location), red line segment tracks the vehicle displacement (design) from the preceding experiment, and contours plot the posterior PDF. }
  \label{fig:Diffusion_vsgreedy}
\end{figure}

With the insight that corners are good design locations,
understanding the behavior of PG-sOED
becomes easier. 
\Cref{fig:Diffusion_vsgreedy_soed1} displays the posterior contours after 1 and 2 experiments (i.e. $p(\theta|I_1)$ and $p(\theta|I_2)$; also recall the prior $p(\theta|I_0)$ is uniform) of an episode instance when following PG-sOED; \cref{fig:Diffusion_vsgreedy_greedy1} displays those for greedy design. In each plot, the purple star represents the true source location for that episode, the red dot represents the physical state (vehicle location), and the red line segment tracks the vehicle displacement (design) from the preceding experiment. 

In PG-sOED (\cref{fig:Diffusion_vsgreedy_soed1}), the first design moves the vehicle towards a corner despite the source is off at $t_0$ and that no concentration signal is obtained, incurring a negative reward $g_0=-0.040$ due to the movement penalty. The greedy design realizes the source is off and remains at the initial location (center), keeping its reward at $g_0=0$.
At this point, it would appear greedy is performing better. The source then becomes active in the second experiment at $t_1$, and both PG-sOED and greedy shift the vehicle towards a corner. However, PG-sOED is able to arrive much closer to the corner and obtains a more informative measurement compared to greedy design, since PG-sOED has already made a head start in the first experiment. Therefore, PG-sOED is able to look ahead and take into account future outcomes. With an initial ``sacrifice'' of seemingly fruitless first experiment, PG-sOED is able to better position the vehicle for a much more lucrative second experiment, such that the expected \emph{total} reward is maximized ($\sum_{k=0}^2 g_k=2.941$ for PG-sOED versus $\sum_
  {k=0}^2 g_k=2.022$ for greedy). 

\begin{figure}[htbp]
  \centering
  \includegraphics[width=0.7\linewidth]{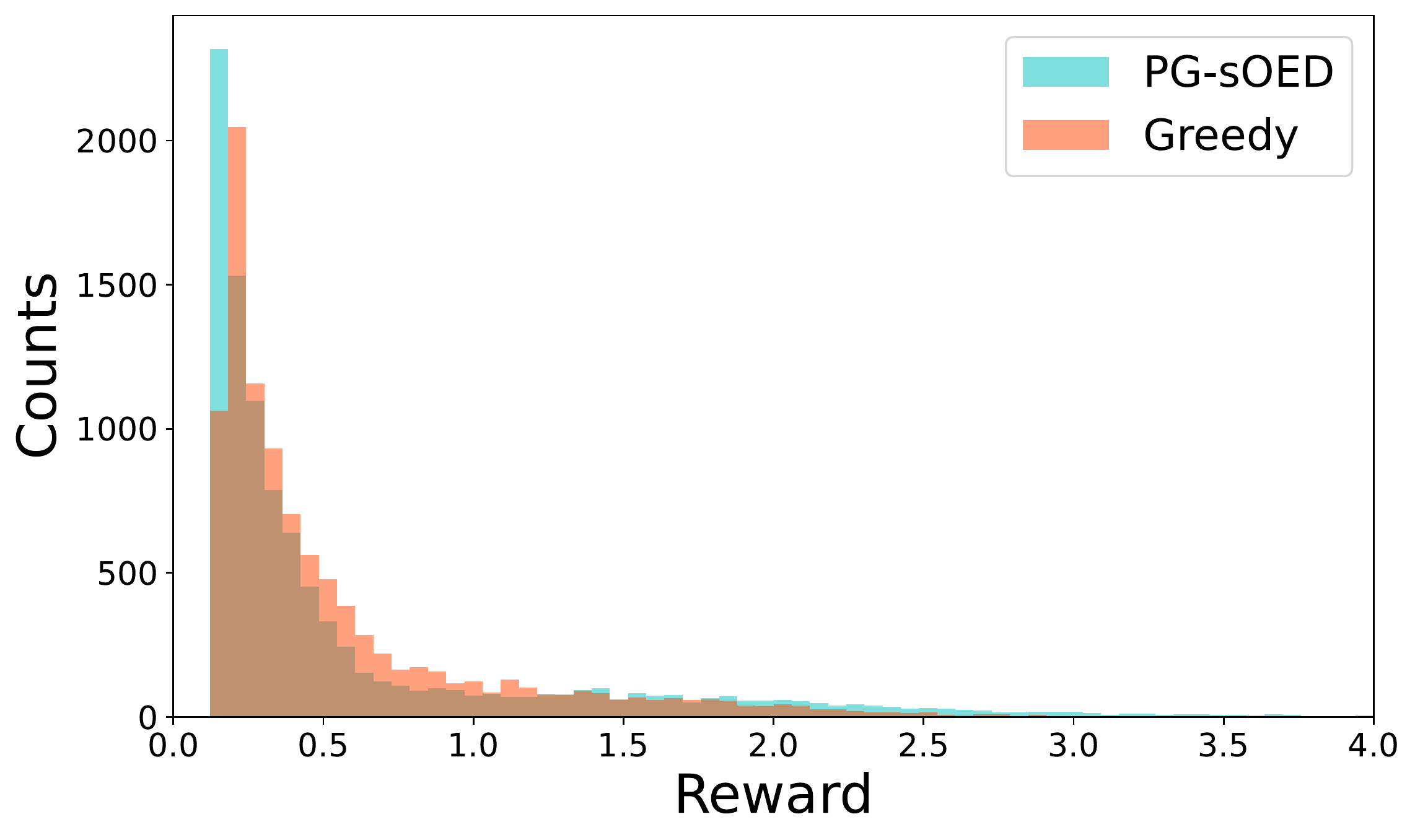}
  \caption{Case 1. Histograms of total rewards from $10^4$ episodes generated using PG-sOED and greedy designs. 
  The mean total reward for PG-sOED is $0.615 \pm 0.007$, higher than greedy design's $0.552 \pm 0.005$.
  }
  \label{fig:Diffusion_hist}
\end{figure}

We further generate $10^4$ episodes under different 
$\theta$ samples drawn from the prior, and collect their realized total rewards in
\cref{fig:Diffusion_hist}. Indeed, the mean total reward for PG-sOED is $0.615 \pm 0.007$, higher than greedy design's $0.552 \pm 0.005$. Note that while PG-sOED has more low-reward episodes {corresponding to when the true source location is far away from the top right corner and when incurring high movement penalty in the first stage}, it also has more high-reward episodes {corresponding to when the true source location is near the top right corner}. Overall, PG-sOED achieves a greater mean value than greedy design. 

\subsubsection{Case 2}
\begin{figure}[htbp]
  \centering
  \includegraphics[width=0.8\linewidth]{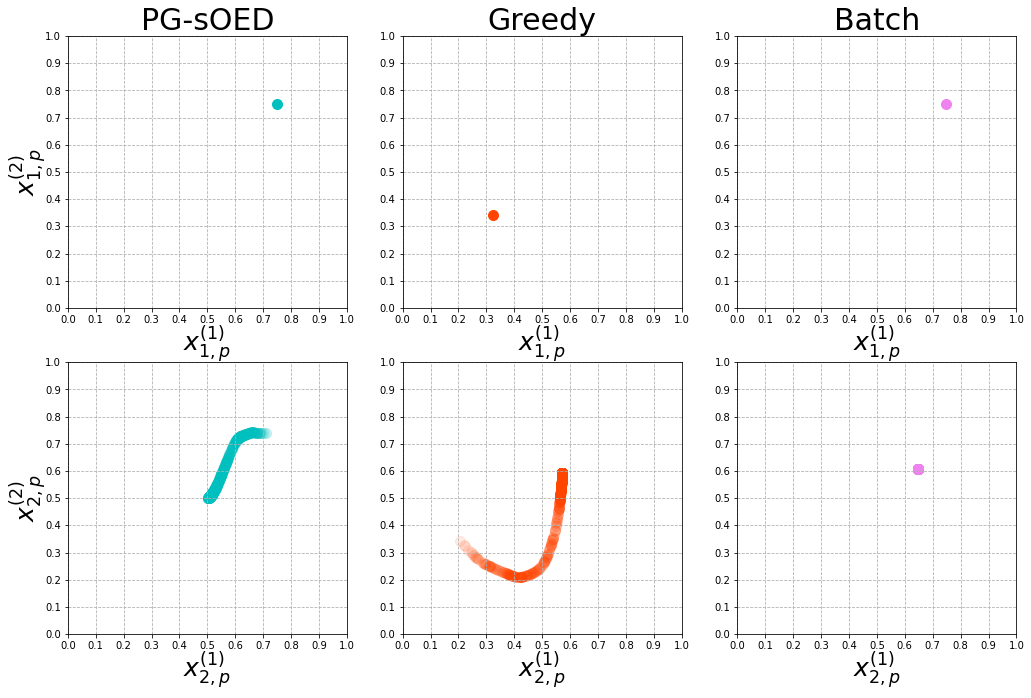}
  \caption{Case 2. Vehicle locations of episodes obtained from PG-sOED, greedy, and batch designs.}
  \label{fig:Source_xp}
\end{figure}

\begin{figure}[htbp]
  \centering
  \subfloat[   $t=0.05$]{\label{fig:Source_util_1}\includegraphics[width=0.48\linewidth]{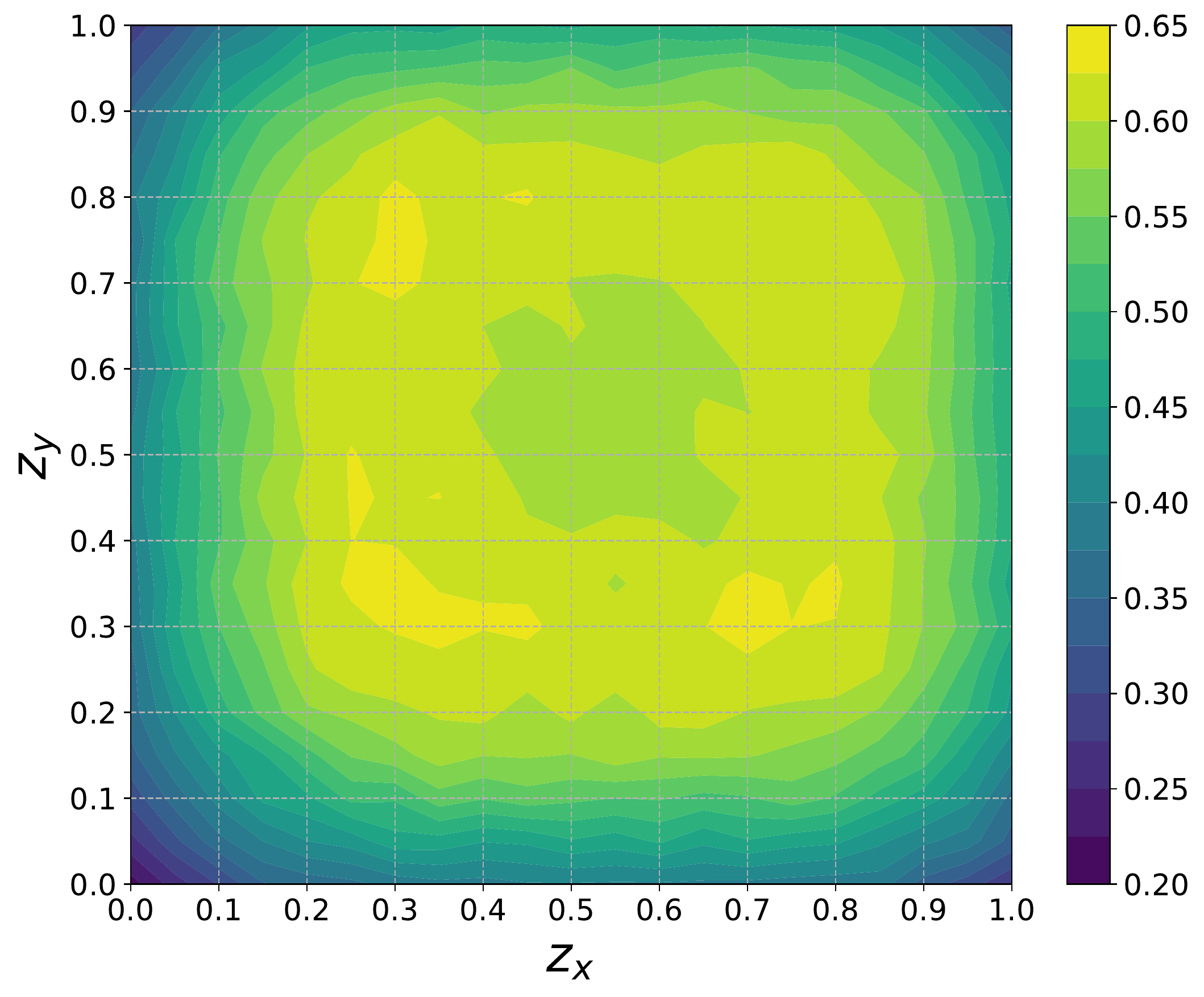}}
  \subfloat[   $t=0.2$]{\label{fig:Source_util_2}\includegraphics[width=0.48\linewidth]{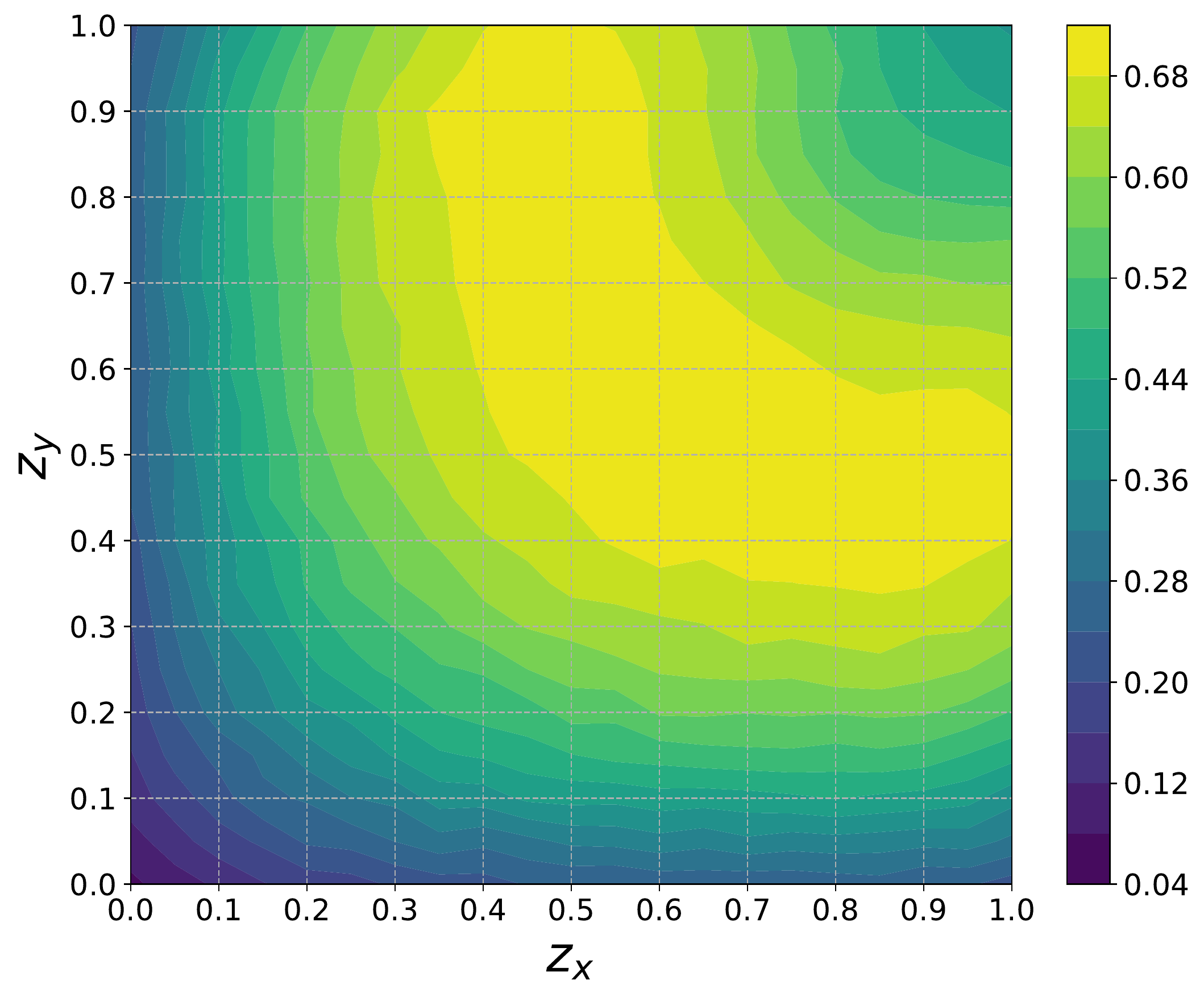}}
  \caption{Case 2. Expected utility versus sensor location if conducting a single experiment at $t=0.05$ or $t=0.2$.
    }
  \label{fig:Source_util}
\end{figure}

Case 2 incorporates convection in addition to diffusion, and disables movement penalty for the vehicle (i.e. $c_q=0$).
In \cref{fig:Source_xp}, we plot the physical states $x_{1,p}$ and $x_{2,p}$ (i.e. vehicle locations after the first and second experiments) from $10^4$ episodes sampled from PG-sOED, greedy, and batch designs. 
We observe both PG-sOED and batch design initially move the vehicle towards the top right corner and then turn back; greedy design roughly moves in the opposite direction. Notably, $d_1$ (design for the second experiment) for batch design is always the same, in contrast to PG-sOED and greedy that are adaptive. 
The behavior of the different policies can be better understood through \cref{fig:Source_util}, which shows the contours of expected utility versus sensor location if performing only a \emph{single} experiment at $t_0$ or $t_1$, respectively. 
In \cref{fig:Source_util_1}, we find the global maximum to be around $(0.3,0.3)$, which explains the initial movement of greedy design towards the bottom left.
However, \cref{fig:Source_util_2} reveals that the top right region becomes more informative at $t_1$. Physically, this makes sense since the convection velocity grows over time towards the top-right direction, and more information can be gathered if we ``catch'' the flow at a downstream position. 
This explains why PG-sOED and batch design both move towards the top right even in the first experiment since both of those designs can see the more informative second experiment (except that batch design cannot adapt). 

\begin{figure}[htbp]
  \centering
  \subfloat[sOED versus greedy]{\label{fig:Source_soed_greedy_hist}\includegraphics[width=0.48\linewidth]{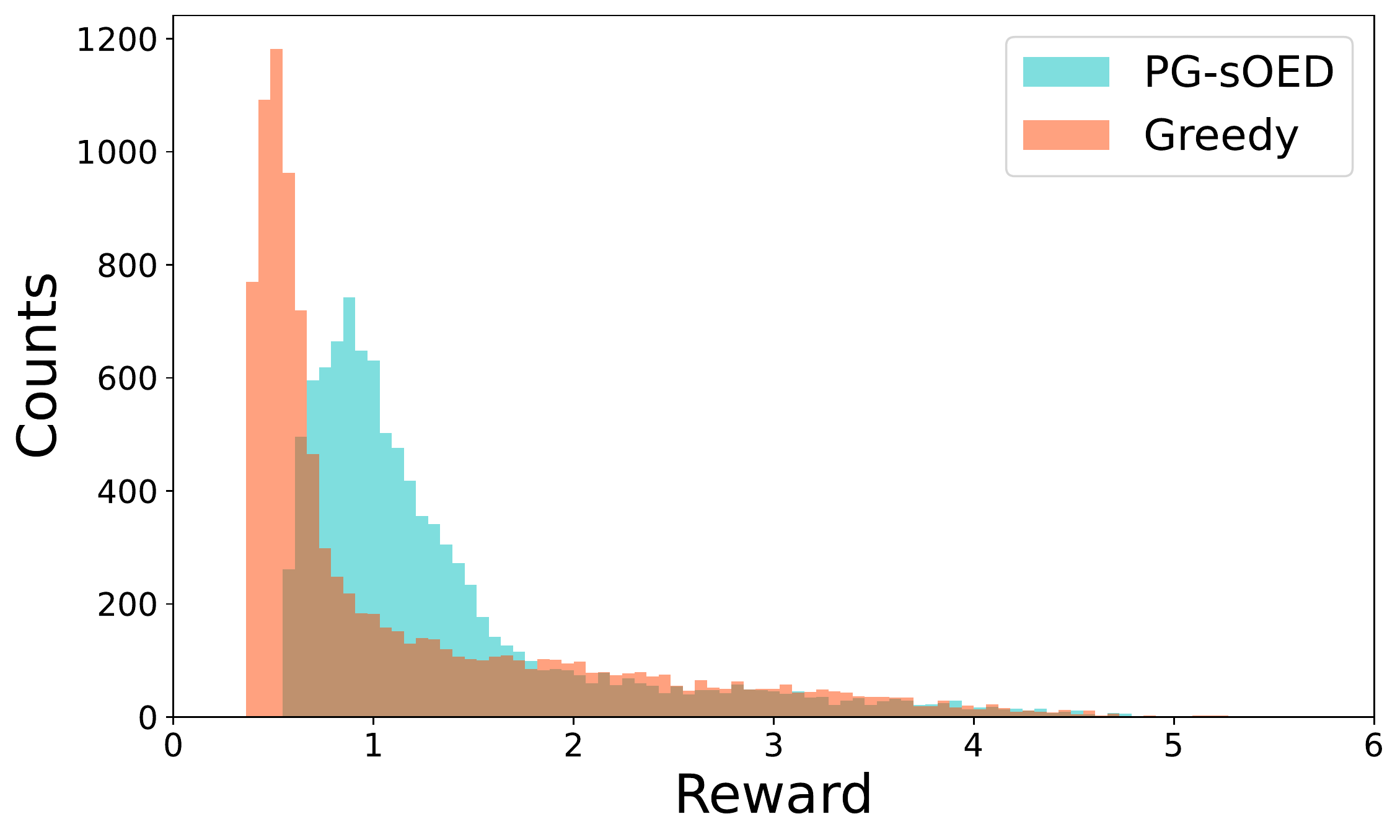}}
  \subfloat[sOED versus batch]{\label{fig:Source_soed_batch_hist}\includegraphics[width=0.48\linewidth]{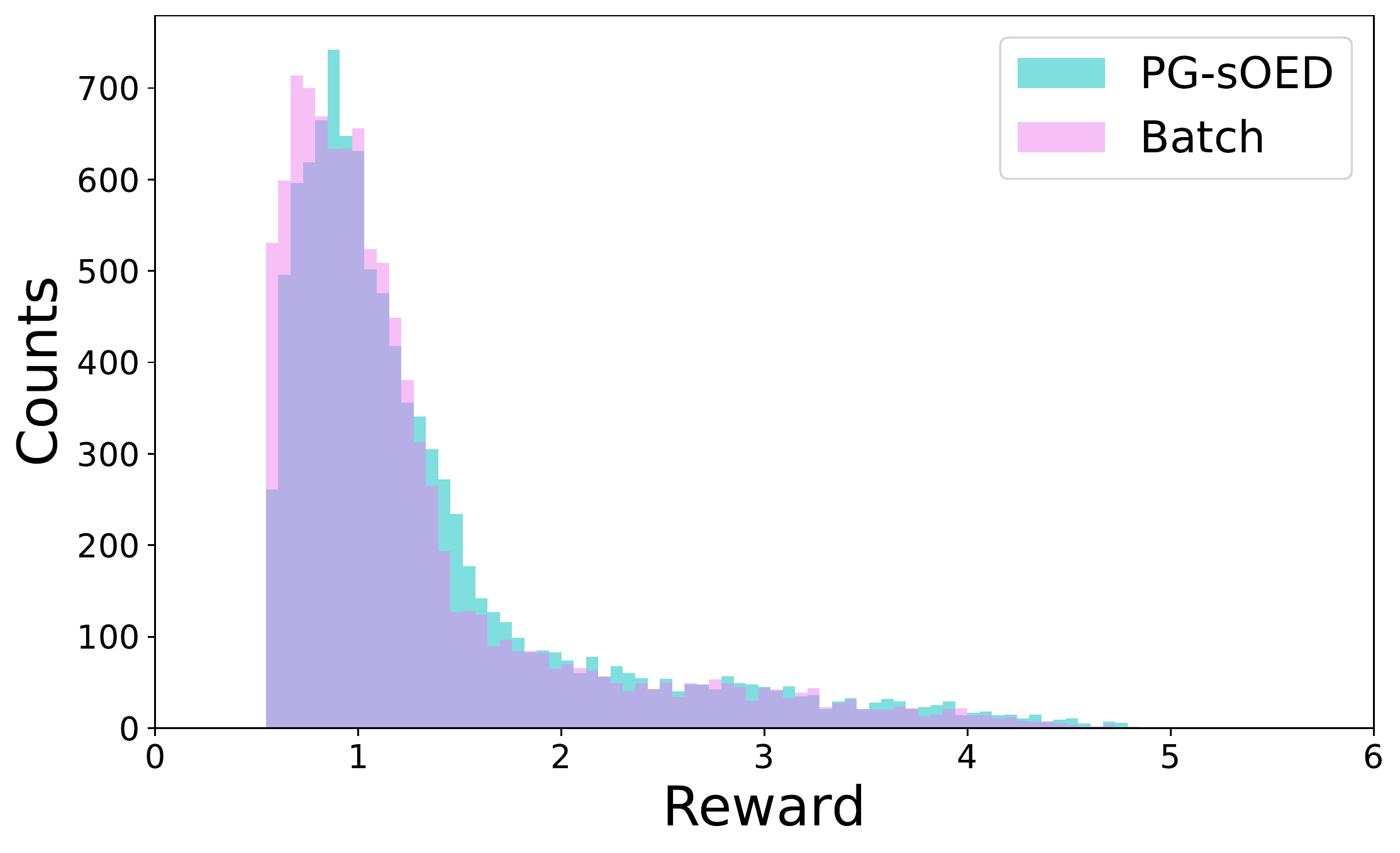}}
  \caption{Case 2. Histograms of total rewards from $10^4$ episodes generated using PG-sOED, greedy, and batch designs. The mean total reward for PG-sOED is $1.344 \pm 0.008$, higher than greedy design's $1.178 \pm 0.010$ and batch design's $1.264 \pm 0.007$.
  }
  \label{fig:Source_hist}
\end{figure}

Back to the two-experiment design, \cref{fig:Source_hist} summarizes the total rewards from all $10^4$ episodes, with PG-sOED having the highest mean value at
$1.344 \pm 0.008$ followed by batch design's $1.264 \pm 0.007$ and greedy design's $1.178 \pm 0.010$.
The advantage of PG-sOED is greater over greedy and less over batch, suggesting a more prominent role of lookahead.
From the histograms, greedy design has many low-reward episodes, corresponding to scenarios when the true source location is in the upper-right region. At the same time, greedy also has a similar distribution of high-reward episodes as sOED because it is able to adapt. In contrast, batch design does not have many low-reward episodes since it moves towards the upper-right in the first experiment. However, it also has fewer high-reward episodes compared to sOED because it is unable to adapt.

\begin{figure}[htbp]
  \centering
  \subfloat[PG-sOED, $\theta=(0.7,0.9)$, total reward $=2.020$]{\label{fig:Source_vsgreedy_soed1}\includegraphics[width=0.48\linewidth]{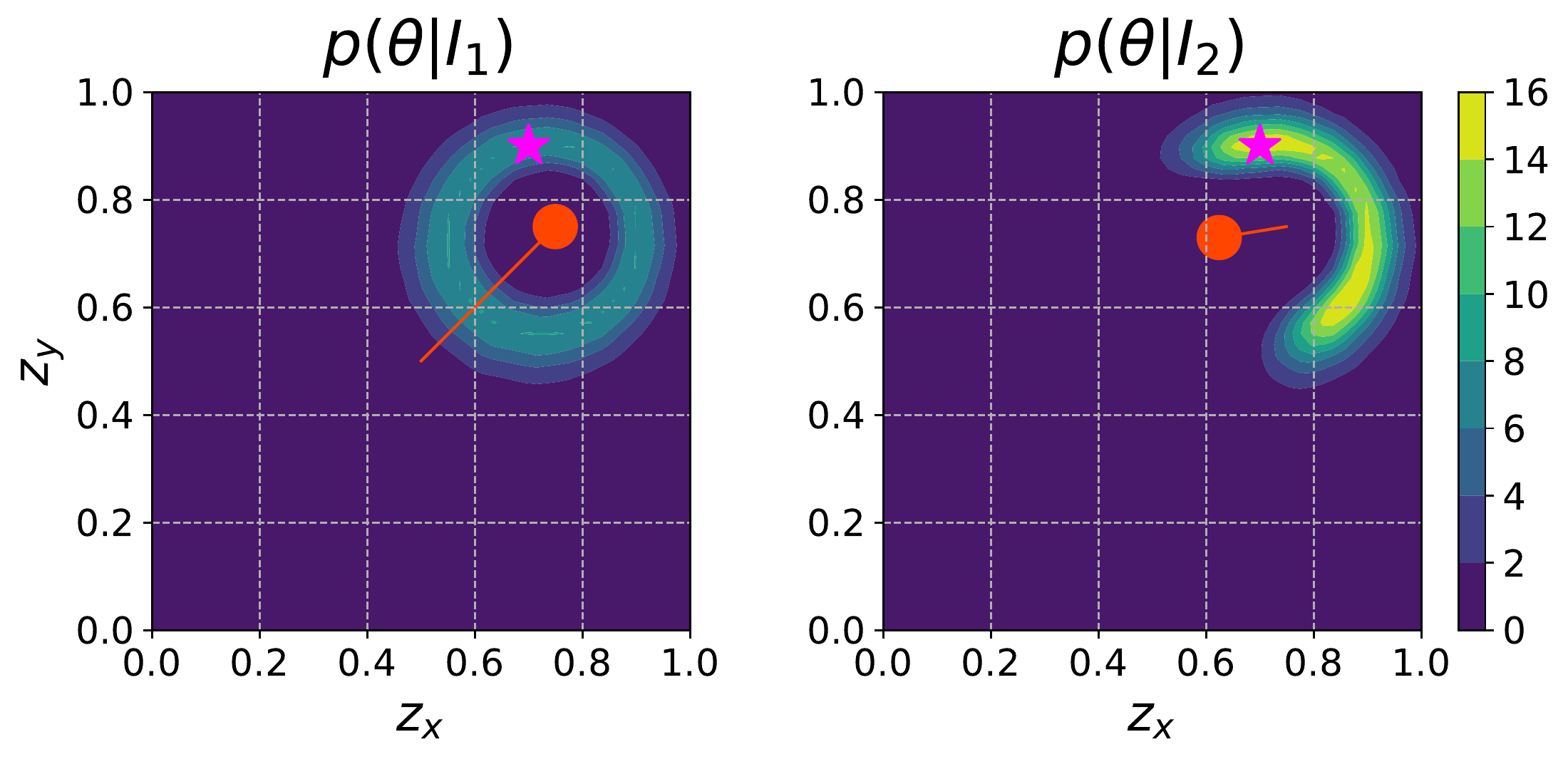}} 
  \subfloat[Greedy, $\theta=(0.7,0.9)$, total reward $=0.493$]{\label{fig:Source_vsgreedy_greedy1}\includegraphics[width=0.48\linewidth]{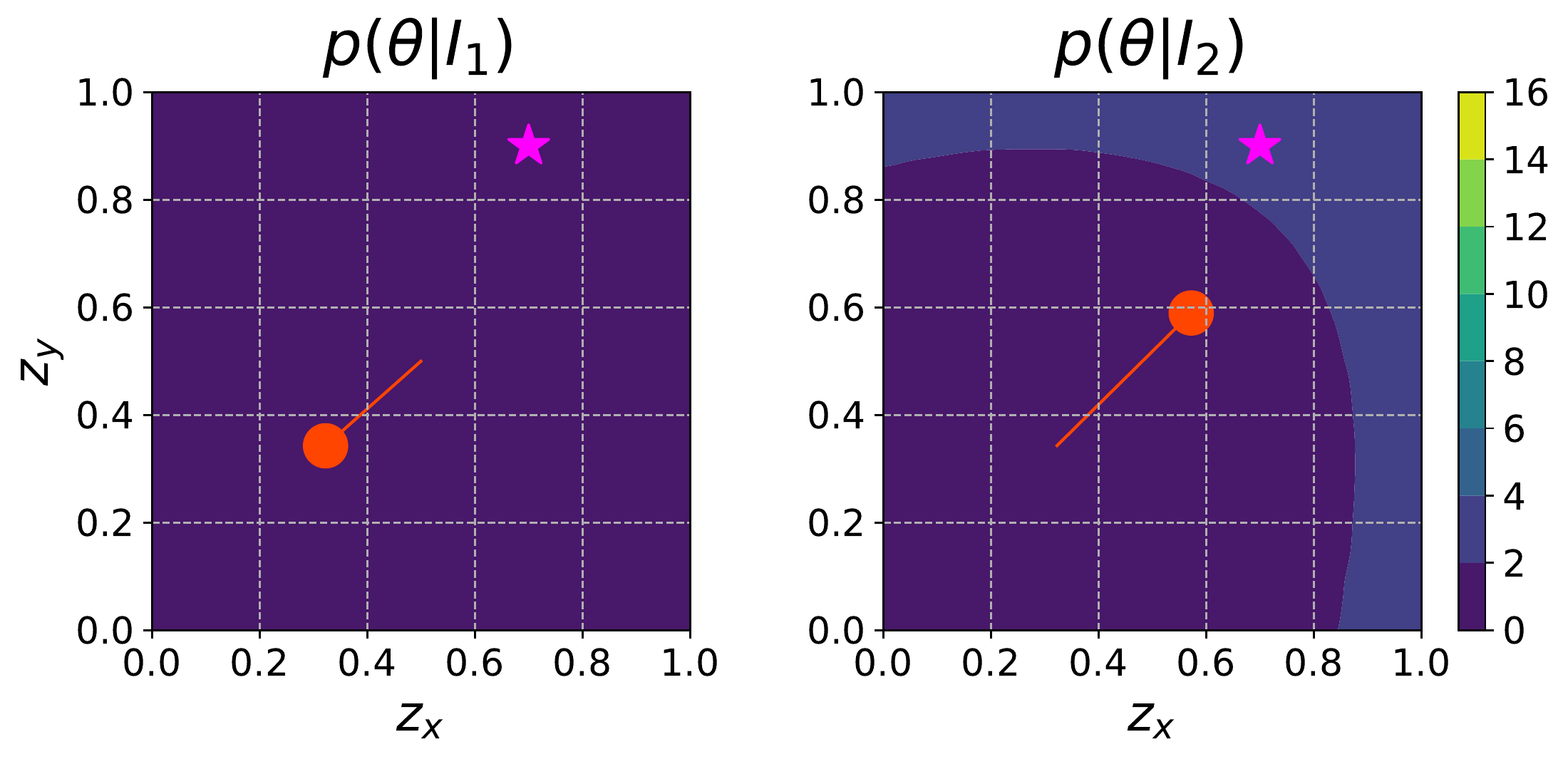}}\\
  \subfloat[PG-sOED, $\theta=(0.65,0.7)$, total reward $=4.014$]{\label{fig:Source_vsbatch_soed}\includegraphics[width=0.48\linewidth]{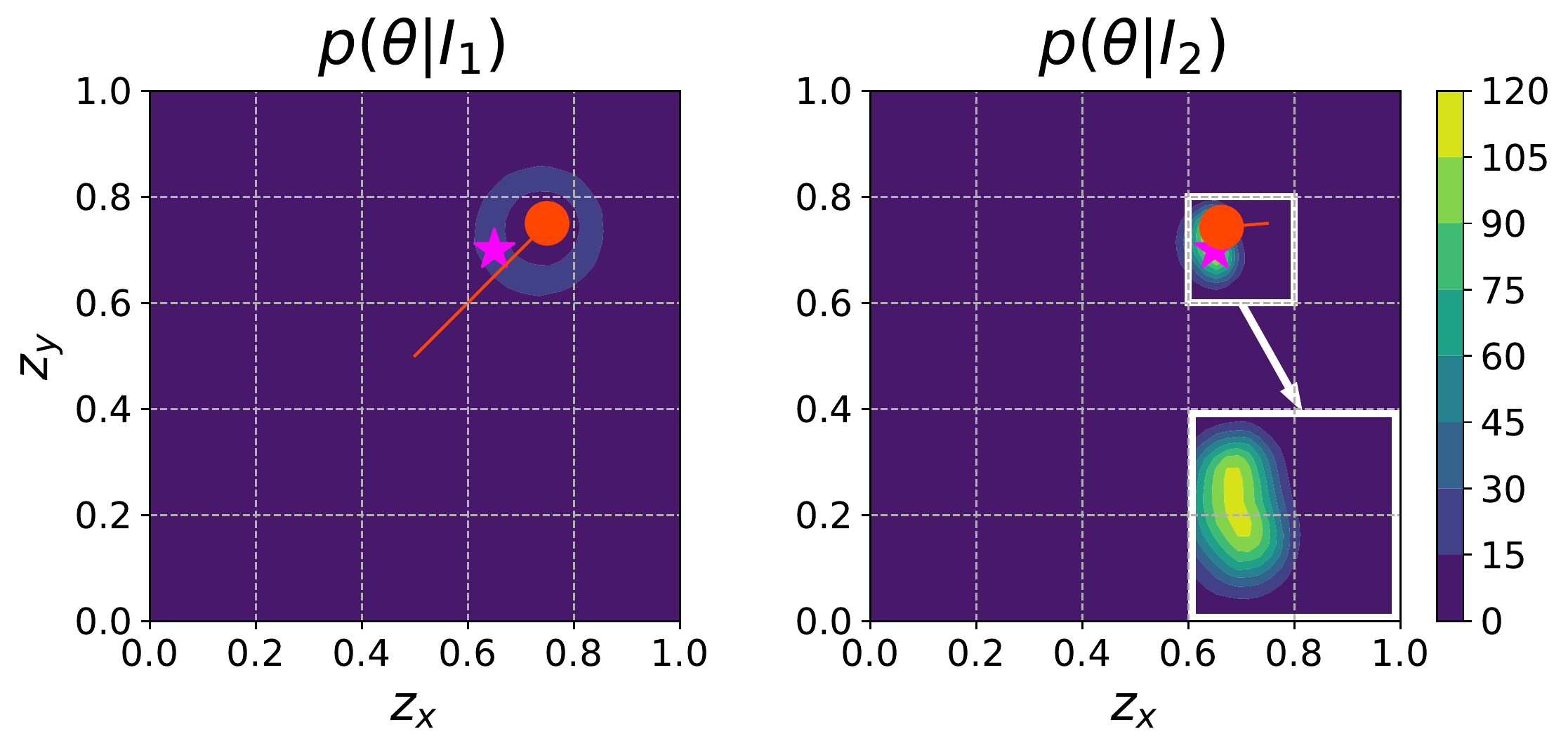}} 
  \subfloat[Batch, $\theta=(0.65,0.7)$, total reward $=2.647$]{\label{fig:Source_vsbatch_batch}\includegraphics[width=0.48\linewidth]{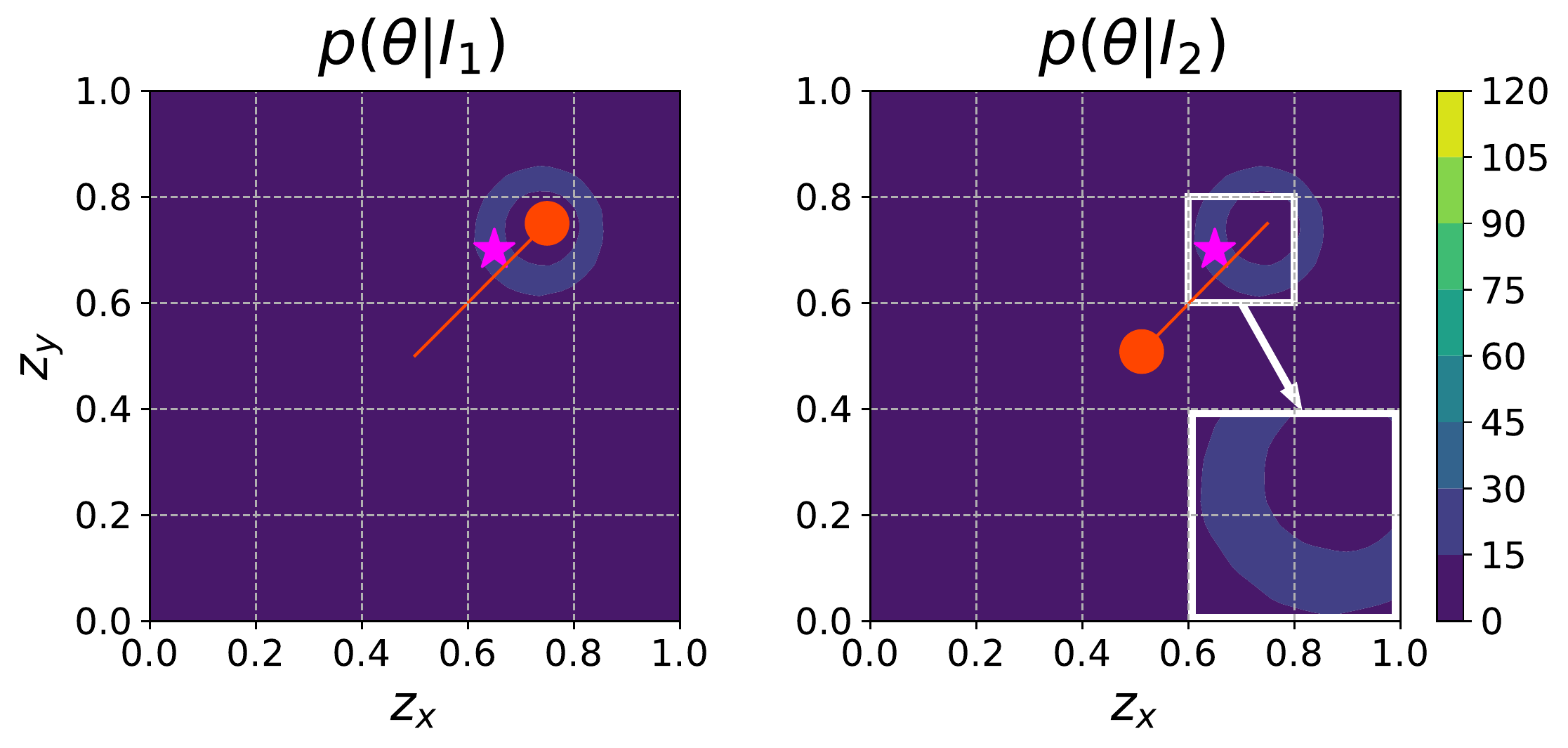}}
\\  \caption{Case 2. 
  Episode instances obtained by PG-sOED, greedy, and batch designs, where PG-sOED outperforms both greedy and batch designs. The purple star represents the true $\theta$, red dot represents the physical state (vehicle location), red line segment tracks the vehicle displacement (design) from the preceding experiment, and contours plot the posterior PDF. The inset zooms in on the high-probability posterior region.
  }
  \label{fig:Source_vsgreedy}
\end{figure}

\begin{figure}[htbp]
  \centering
  \subfloat[PG-sOED, $\theta=(0.1,0.4)$, total reward $=1.076$]{\label{fig:Source_vsgreedy_soed2}\includegraphics[width=0.48\linewidth]{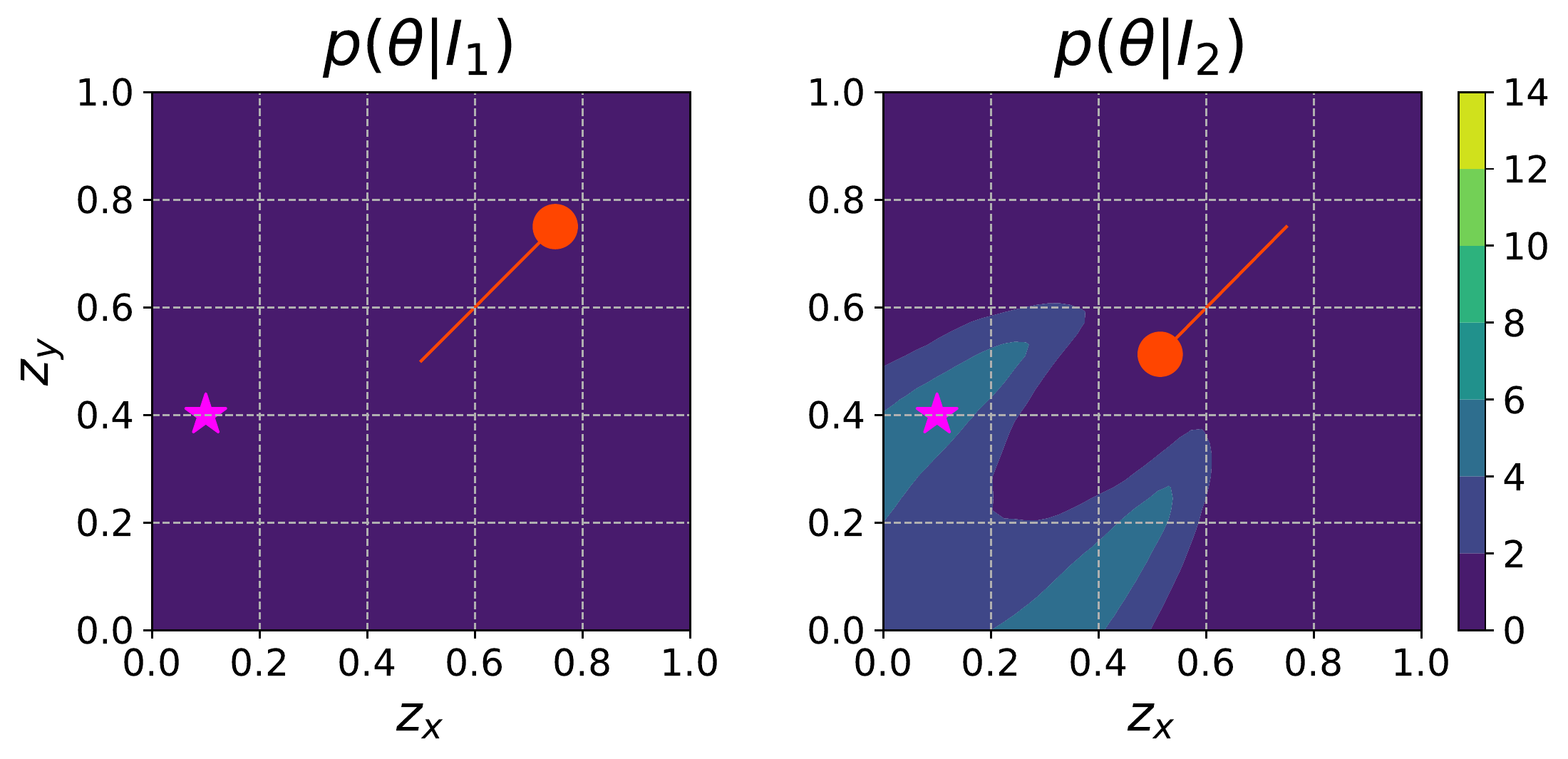}} 
  \subfloat[Greedy, $\theta=(0.1,0.4)$, total reward $=1.687$]{\label{fig:Source_vsgreedy_greedy2}\includegraphics[width=0.48\linewidth]{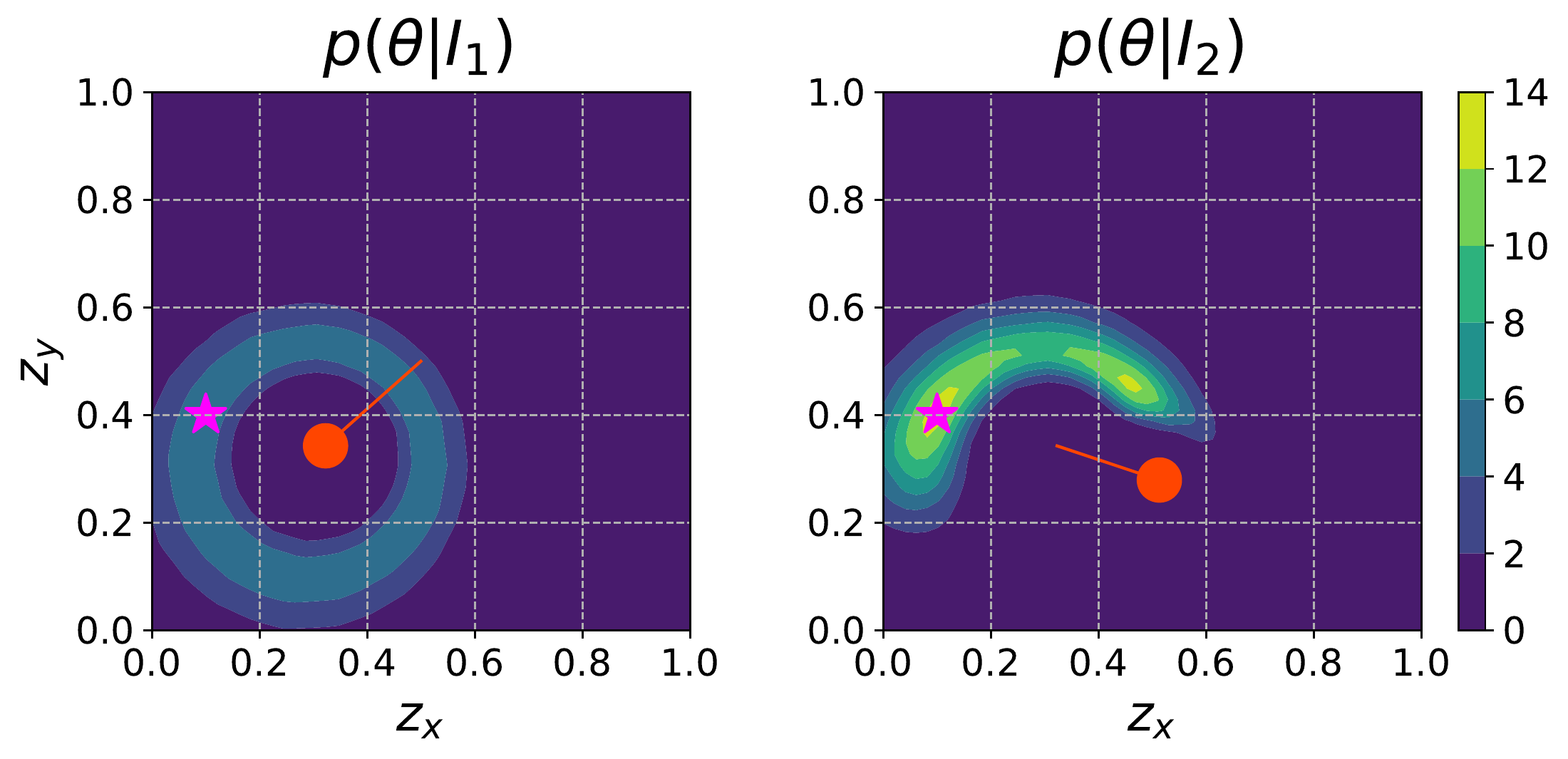}}
  \caption{Case 2. 
  Episode instances obtained by PG-sOED and greedy design, where greedy design outperforms PG-sOED. The purple star represents the true $\theta$, red dot represents the physical state (vehicle location), red line segment tracks the vehicle displacement (design) from the preceding experiment, and contours plot the posterior PDF.}
  \label{fig:Source_vsbatch}
\end{figure}

Lastly, we provide examples of posteriors resulting from sample episodes. \Cref{fig:Source_vsgreedy} presents scenarios where PG-sOED visibly achieves a ``narrower'' posterior compared to greedy and batch designs, which is reflected quantitatively through the higher total reward. However, there are also scenarios 
where PG-sOED achieves a lower total reward, such as shown in \cref{fig:Source_vsbatch}. It is the \emph{expected} utility averaged over all possible scenarios that PG-sOED maximizes.

\subsubsection{Case 3}

Case 3 provides demonstration on a problem with longer design horizon ($N=4$ experiments) and a higher dimensional parameter space ($N_\theta=4$) that now also includes the source strength $\theta_s$ and width $\theta_h$. 

\begin{figure}[htbp]
  \centering
  \subfloat[PG-sOED versus greedy]{\label{fig:Source3_soed_greedy_hist}\includegraphics[width=0.48\linewidth]{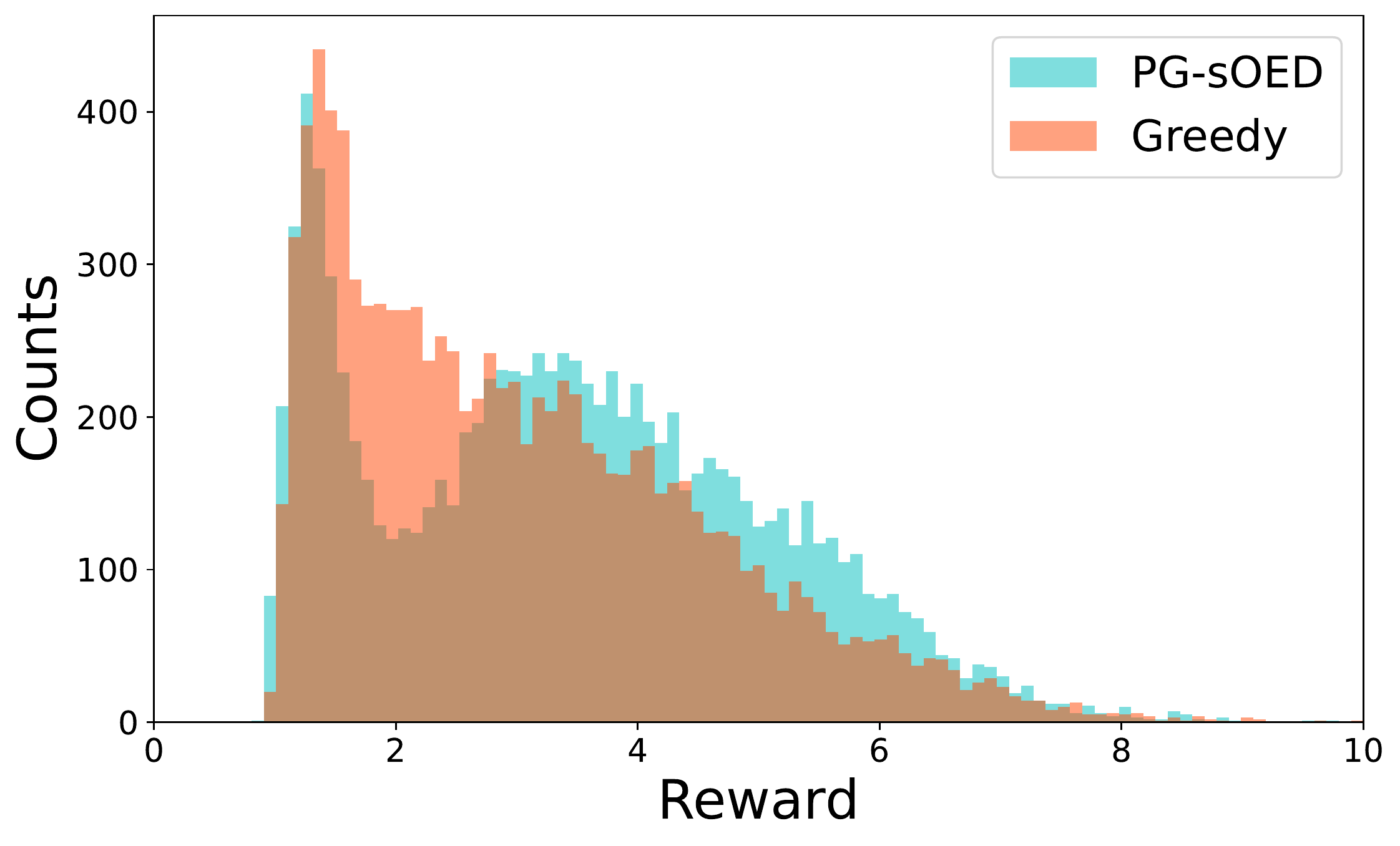}}
  \subfloat[PG-sOED versus batch]{\label{fig:Source3_soed_batch_hist}\includegraphics[width=0.48\linewidth]{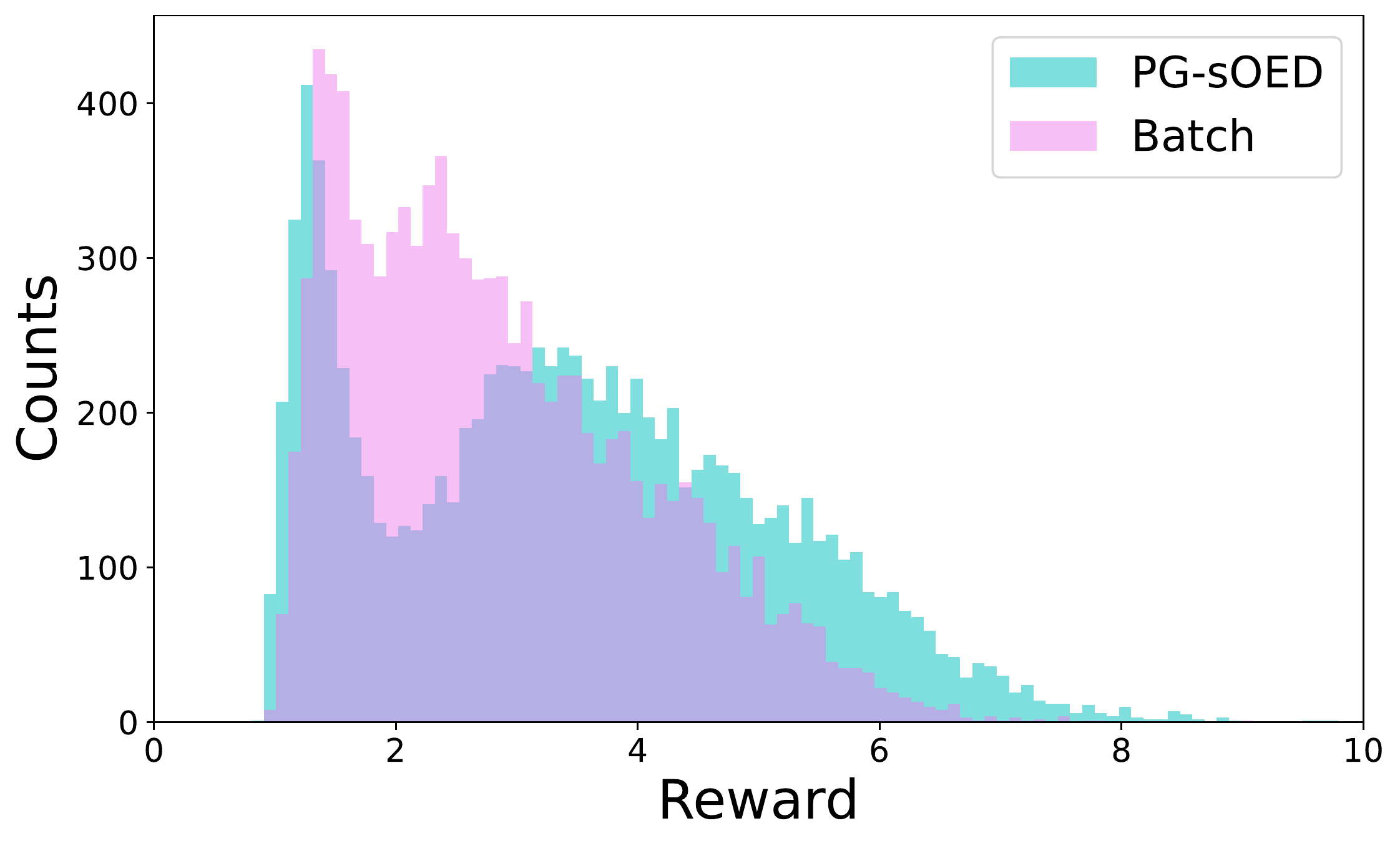}}
  \caption{Case 3. Histograms of total rewards from $10^4$ episodes generated using PG-sOED, greedy, and batch designs. The mean total reward for PG-sOED is $3.435 \pm 0.016$, higher than greedy design's $3.057 \pm 0.015$ and batch design's $2.856 \pm 0.012$.
}
  \label{fig:Source3_hist}
\end{figure}

\begin{figure}[htbp]
  \centering
  \includegraphics[width=0.85\linewidth]{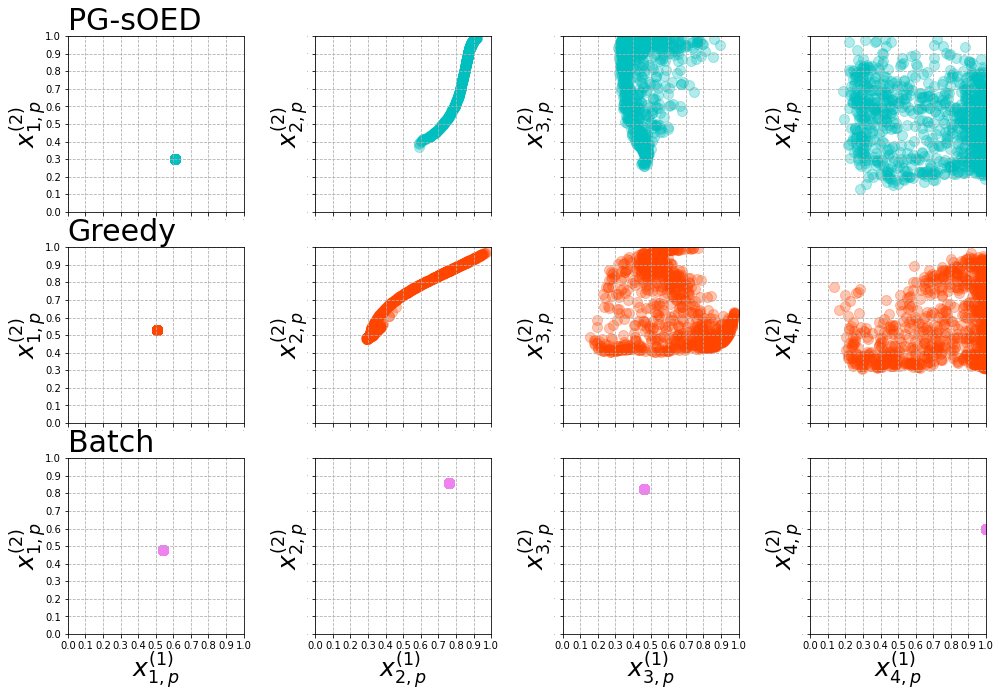}
  \caption{Case 3. Vehicle locations of episodes obtained from PG-sOED, greedy, and batch designs (rows) plotted for experiments 1--4 (columns).}
  \label{fig:Source3_xp}
\end{figure}

From \cref{fig:Source3_hist}, we see that PG-sOED's mean total reward ($3.435 \pm 0.016$) outperforms both greedy ($3.057 \pm 0.015$) and batch ($2.856 \pm 0.012$) designs.
In particular, PG-sOED features a prominent bimodal distribution of the total rewards, but also a heavier tail to the right leading to an overall greater total reward \emph{in expectation} compared to greedy and batch designs. 
In \cref{fig:Source3_xp}, we plot the physical states $x_{k+1,p}$ (i.e. vehicle locations after the $k$-th experiments) from $10^4$ episodes sampled from PG-sOED, greedy, and batch designs. As expected, batch design always produces identical movement paths since it is non-adaptive, while the PG-sOED and greedy designs can spread into the entire domain. Notably, we do not observe movement to the left and bottom regions even after all four experiments. This is because moving towards the bottom left is against the wind direction and requires a higher movement cost, and the bottom left does not offer measurements as informative as the top right since the convection direction is towards the top right.

\begin{figure}[htbp]
  \centering
  \subfloat[PG-sOED, $\theta=(0.6,0.7,0.06,0.2)$, total reward $=3.428$]{\label{fig:Source3_contour_soed}\includegraphics[width=0.8\linewidth]{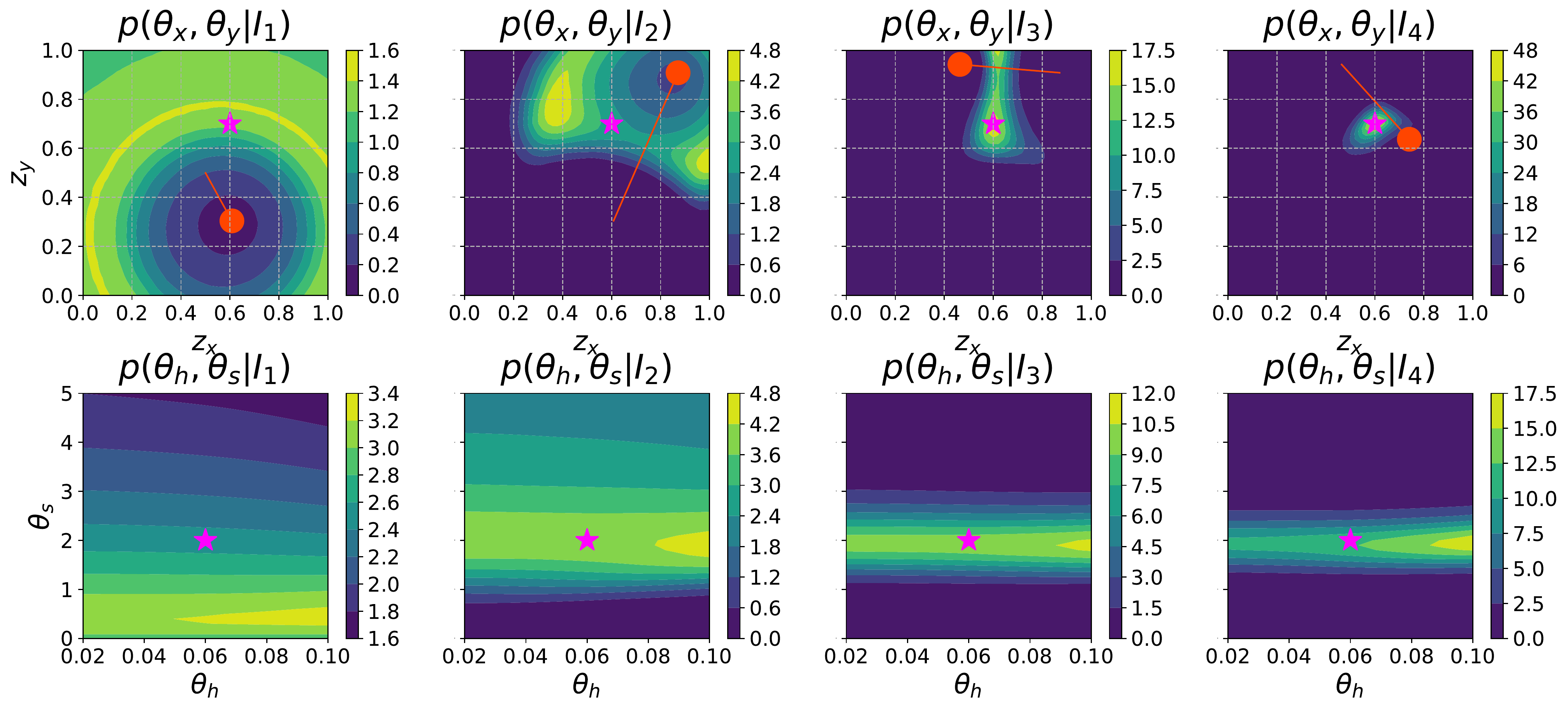}} \\
  \subfloat[Greedy, $\theta=(0.6,0.7,0.06,0.2)$, total reward $=2.614$]{\label{fig:Source3_contour_greedy}\includegraphics[width=0.8\linewidth]{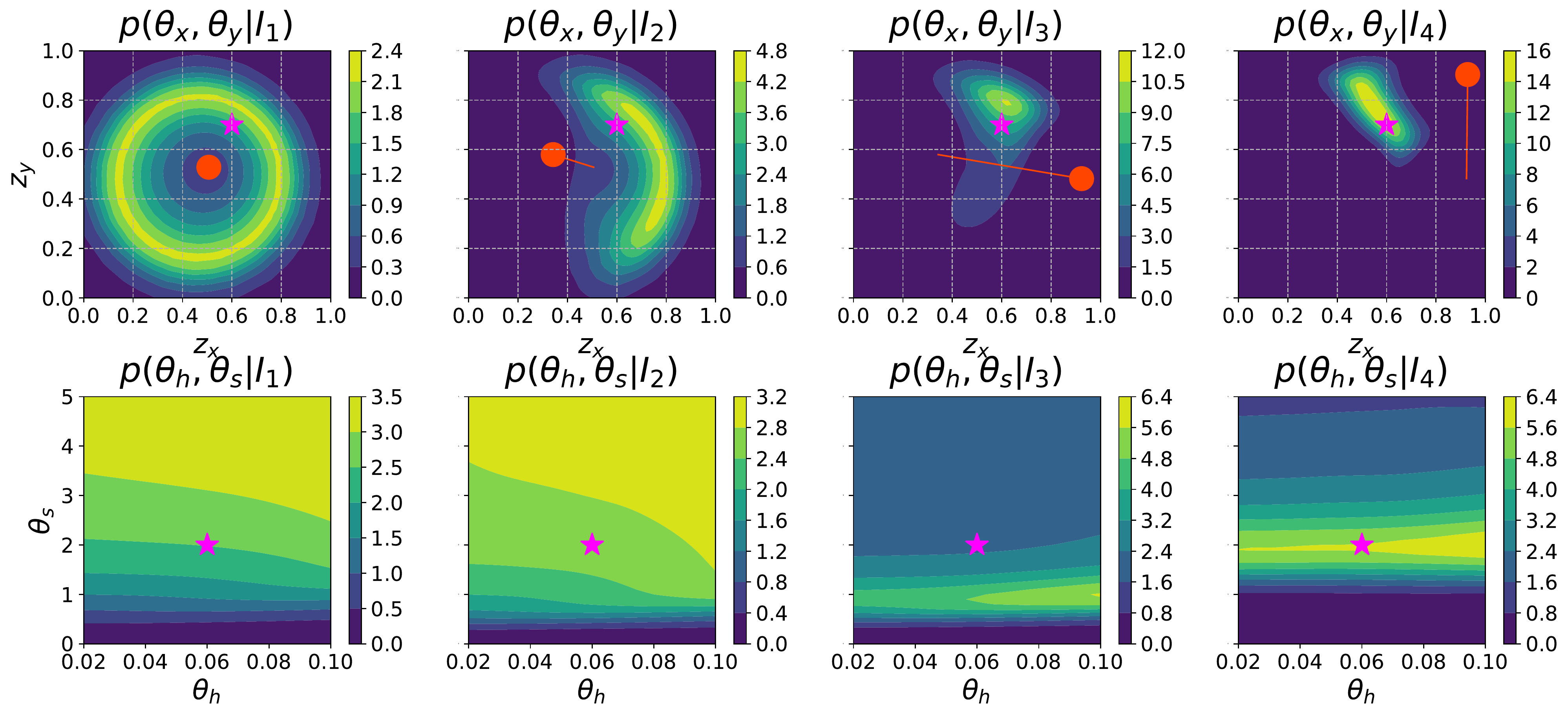}} \\
  \subfloat[Batch, $\theta=(0.6,0.7,0.06,0.2)$, total reward $=1.984$]{\label{fig:Source3_contour_batch}\includegraphics[width=0.8\linewidth]{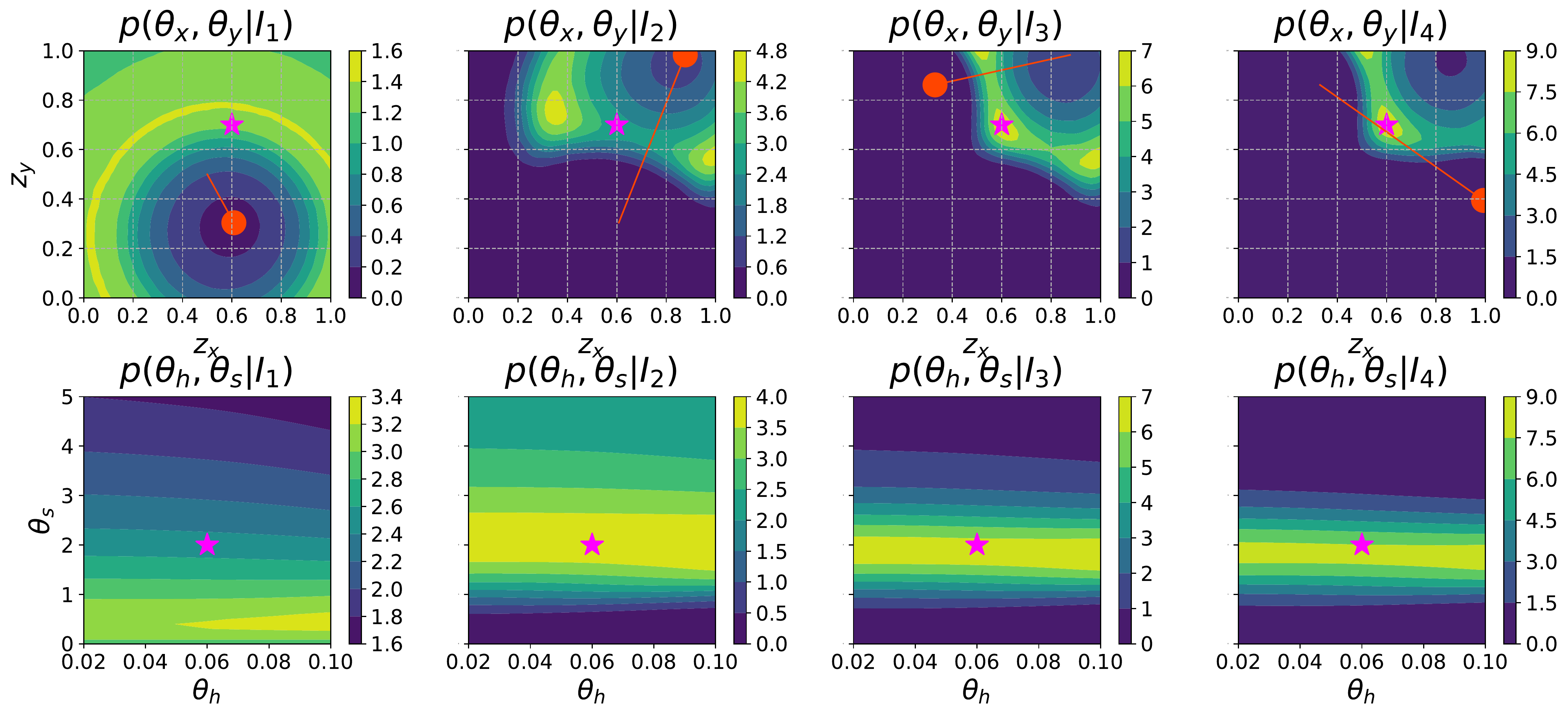}} 
  \caption{Case 3. 
  Episode instances obtained by PG-sOED, greedy and batch designs. The purple star represents the true $\theta$, red dot represents the physical state (vehicle location), red line segment tracks the vehicle displacement (design) from the preceding experiment, and contours plot the marginal posterior PDF.}
  \label{fig:Source3_contour}
\end{figure}

\cref{fig:Source3_contour} shows the marginal posterior PDF contours from episode instances obtained by PG-sOED, greedy and batch designs. Interestingly, both PG-sOED and batch design move towards the bottom right corner in the first experiment, leading to a marginal posterior in $\theta_x$ and $\theta_y$ with a half circle shape that provides a better setup for the next movement. In contrast, the marginal posterior of greedy design leads to a full circle shape, which does not narrow down the direction as much. Consequently, PG-sOED outperforms greedy design. However, this initial advantage in batch design is overtaken by its lack of adaptation in the subsequent experiments, leading to a lower expected total reward than greedy design. We also find the posterior mode 
for $\theta_h$ to be noticeably shifted from its true data-generating value, and the uncertainty remains high even after four experiments. 
This can be explained by the low sensitivity of the concentration to the range of source widths considered, making $\theta_h$ more difficult to identify.
However, the mode of the full four-dimensional joint posterior (not plotted) is still close to the true value.

\section{Conclusions}
\label{sec:conclusions}

This paper presents a mathematical framework and computational methods to optimally design a finite number of sequential experiments (sOED); the code is available at \url{https://github.com/wgshen/sOED}. 
We formulate sOED as a finite-horizon POMDP. 
This sOED form is provably optimal, incorporates both elements of feedback and lookahead, and generalizes the suboptimal batch (static) and greedy (myopic) design strategies. 
We further structure the sOED problem in a fully Bayesian manner and with information-theoretic rewards (utilities), and prove the equivalence of incremental and terminal information gain setups. In particular, sOED can accommodate expensive nonlinear forward models with general non-Gaussian posteriors of continuous random variables.

We then introduce numerical methods for solving the sOED problem, which entails finding the optimal policy that maximizes the expected total reward.
At the core of our approach is PG, an actor-critic RL technique that parameterizes and learns both the policy and value functions in order to extract the gradient with respect to the policy parameters.
We derive and prove the PG expression for finite-horizon sOED, and propose an MC estimator. 
Accessing derivative information enables the use of gradient-based optimization algorithms to achieve  efficient policy search. 
Specifically, we parameterize the policy and value functions as DNNs, and detail architecture choices that accommodate a nonparametric representation of the Bayesian posterior belief states. Further combined with a terminal information gain formulation, the Bayesian inference becomes embedded in the design sequence, allowing us to sidestep the need for explicitly and numerically computing the Bayesian posteriors at intermediate experiments.

We apply the overall PG-sOED method to two different examples.
The first is a linear-Gaussian problem that offers a closed form solution, serving as a benchmark. We validate the PG-sOED policy against the analytic optimal policy, and observe orders-of-magnitude speedups of PG-sOED over an ADP-sOED baseline.
The second entails a problem of contaminant source inversion in a convection-diffusion field. Through multiple sub-cases, we illustrate the advantages of PG-sOED over greedy and batch designs, and provide insights to the value of feedback and lookahead in the context of time-dependent convection-diffusion processes. 
This demonstration also illustrates the ability of PG-sOED to accommodate expensive forward models with nonlinear physics and dynamics.

The main limitation of the current PG-sOED method is its inability 
to handle high-dimensional settings. While the nonparametric representation sidesteps the need to compute intermediate posteriors, Bayesian inference is ultimately required in order to estimate the KL divergence in the terminal reward.
Thus, an important direction of future work is to improve scalability for high-dimensional inference, to go beyond the current gridding method. This may be approached by employing more general and approximate inference methods such as MCMC, variational inference, approximate Bayesian computation, and transport maps,  perhaps in combination with dimension-reduction techniques.

Another fruitful area to explore is within advanced RL techniques
(e.g., \cite{mnih2015human,lillicrap2015continuous,mnih2013playing, 
schulman2017proximal}).
For example, replay buffer stores the experienced episodes, and training data can be sampled from this buffer to reduce sampling costs, control correlation among samples, and reach better convergence performance. 
Off-policy algorithms track two version of the policy network and Q-network---a behavior network for determining actions and a target network for learning---which have demonstrated improved sample efficiency.
Parameters of the policy and Q-networks may also be shared due to their similar features.
Finally, adopting new utility measures, such as those reflecting goal-orientedness, robustness, and risk, would be of great interest to better capture the value of experiments and data in real-life and practical settings.

\appendix

\section{Equivalence of Incremental and Terminal Information Gain in sOED} 
\label{app:incre_terminal}

\begin{proof}[Proof of \cref{prop:terminal_incremental}]
Upon substituting \cref{eq:terminal1,eq:terminal_info_gN} into  \cref{eq:expected_utility}, the expected utility for a given deterministic policy $\pi$ using the terminal formulation is
\begin{align}
    U_T(\pi)&=\EE_{y_0,...,y_{N-1}|\pi,x_0}\[\int_{\Theta} p(\theta|I_N)\ln{\frac{p(\theta|I_N)}{p(\theta|I_0)}}\,d\theta \] \nonumber\\
    &= \EE_{I_1,\dots,I_N|\pi,x_0}\[\int_{\Theta} p(\theta|I_N)\ln{\frac{p(\theta|I_N)}{p(\theta|I_0)}}\,d\theta \]
    \label{e:app_UT}
\end{align}
where recall $I_k=\{ d_0,y_0,\dots,d_{k-1},y_{k-1} \}$ (and $I_0=\emptyset$).
Similarly, substituting \cref{eq:incremental1,eq:incremental2}, the expected utility for the same policy $\pi$ using the incremental formulation is
\begin{align}
    U_I(\pi)&=\EE_{y_0,...,y_{N-1}|\pi,x_0}\[\sum_{k=1}^N \int_{\Theta} p(\theta|I_k)\ln{\frac{p(\theta|I_k)}{p(\theta|I_{k-1})}}\,d\theta \] \nonumber\\
    &=\EE_{I_1,\dots,I_N|\pi,x_0}\[\sum_{k=1}^N \int_{\Theta} p(\theta|I_k)\ln{\frac{p(\theta|I_k)}{p(\theta|I_{k-1})}}\,d\theta \].
    \label{e:app_UI}
\end{align}
In both cases, 
$\EE_{y_0,\dots,y_{N-1}|\pi,x_0}$ can be equivalently replaced by $\EE_{I_1,\dots,I_N|\pi,x_0}$ since
\begin{align*}
    \EE_{I_1,\dots,I_N |\pi,x_0} \[\cdots\] &= \EE_{d_0,y_0,d_1,y_1,\dots,d_{N-1},y_{N-1} | \pi,x_0} \[\cdots\] \\
    &= \EE_{d_0|\pi} \EE_{y_0,d_1,y_1,\dots,d_{N-1},y_{N-1}|\pi,x_0,d_0} \[\cdots\] \\
    &= \EE_{y_0,d_1,y_1,\dots,d_{N-1},y_{N-1}|\pi,x_0,\mu_0(x_0)} \[\cdots\] \\
    &= \EE_{y_0,d_1,y_1,\dots,d_{N-1},y_{N-1}|\pi,x_0} \[\cdots\] \\
    &= \EE_{y_0|\pi,x_0} \EE_{d_1|\pi,x_0,y_0} \EE_{y_1,\dots,d_{N-1},y_{N-1}|\pi,x_0,y_0,d_1} \[\cdots\] \\
    &= \EE_{y_0|\pi,x_0} \EE_{y_1,\dots,d_{N-1},y_{N-1}|\pi,x_0,y_0,\mu_1(x_1)} \[\cdots\] \\
    &= \EE_{y_0|\pi,x_0} \EE_{y_1,\dots,d_{N-1},y_{N-1}|\pi,x_0,y_0} \[\cdots\] \\
    &= \EE_{y_0|\pi,x_0} \EE_{y_1|\pi,x_0,y_0} \EE_{d_2,\dots,d_{N-1},y_{N-1}|\pi,x_0,y_0,y_1} \[\cdots\] \\
    & \qquad\vdots \\
    &= \EE_{y_0|\pi,x_0} \EE_{y_1|\pi,x_0,y_0} \cdots \EE_{y_{N-1}|\pi,x_0,y_0,y_1,\dots,y_{N-2},\mu_{N-1}(x_{N-1})} \[\cdots\] \\
    &= \EE_{y_0|\pi,x_0} \EE_{y_1|\pi,x_0,y_0} \cdots \EE_{y_{N-1}|\pi,x_0,y_0,y_1,\dots,y_{N-2}} \[\cdots\] \\
    &= \EE_{y_0,\dots,y_{N-1}|\pi,x_0} \[\cdots\],
\end{align*}
where the third equality is due to the deterministic policy (Dirac delta function) $d_0=\mu_0(x_0)$, the fourth equality is due to 
$\mu_0(x_0)$ being known if $\pi$ and $x_0$ are given. The seventh equality is due to $\mu_1(x_1)$ being known if $\pi$ and $x_1$ are given, and $x_1$ is known if $x_0$, $d_0=\mu_0(x_0)$ and $y_0$ are given, and $\mu_0(x_0)$ is known if $\pi$ and $x_0$ are given, so overall $\mu_1(x_1)$ is known if $\pi$, $x_0$ and $y_0$ are given.
The eighth to second-to-last equalities all apply the same reasoning recursively. The last equality brings the expression back to a conditional joint expectation. 

Taking the difference between \cref{e:app_UT} and \cref{e:app_UI}, we obtain
\begin{align*}
    &U_I(\pi) - U_T(\pi)\\
    &=\EE_{I_1,\dots,I_N|\pi,x_0}\[\sum_{k=1}^N \int_{\Theta} p(\theta|I_k)\ln{\frac{p(\theta|I_k)}{p(\theta|I_{k-1})}}\,d\theta - \int_{\Theta} p(\theta|I_N)\ln{\frac{p(\theta|I_N)}{p(\theta|I_0)}}\,d\theta \]\\
    &=\int_{\Theta}\EE_{I_1,\dots,I_N|\pi,x_0}\[ \sum_{k=1}^N p(\theta|I_k)\ln{\frac{p(\theta|I_k)}{p(\theta|I_{k-1})}} - p(\theta|I_N)\ln{\frac{p(\theta|I_N)}{p(\theta|I_0)}} \]\, d\theta\\
    &=\int_{\Theta}\EE_{I_1,\dots,I_N|\pi,x_0}\[ \sum_{k=1}^{N-1} p(\theta|I_k)\ln{\frac{p(\theta|I_k)}{p(\theta|I_{k-1})}} + p(\theta|I_N)\ln{\frac{p(\theta|I_0)}{p(\theta|I_{N-1})}} \]\,d\theta\\
    &=\int_{\Theta}\EE_{I_1,\dots,I_{N-1}|\pi,x_0} \int_{I_N} p(I_N|I_{N-1},\pi) \[ \sum_{k=1}^{N-1} p(\theta|I_k)\ln{\frac{p(\theta|I_k)}{p(\theta|I_{k-1})}} + p(\theta|I_N)\ln{\frac{p(\theta|I_0)}{p(\theta|I_{N-1})}} \]\,dI_N\,d\theta\\
    &=\int_{\Theta}\EE_{I_1,\dots,I_{N-1}|\pi,x_0} \[ \sum_{k=1}^{N-1} p(\theta|I_k)\ln{\frac{p(\theta|I_k)}{p(\theta|I_{k-1})}} + \int_{I_N} p(\theta,I_N|I_{N-1},\pi)\ln{\frac{p(\theta|I_0)}{p(\theta|I_{N-1})}}\,dI_N \]\,d\theta\\
    &=\int_{\Theta}\EE_{I_1,\dots,I_{N-1}|\pi,x_0} \[ \sum_{k=1}^{N-1} p(\theta|I_k)\ln{\frac{p(\theta|I_k)}{p(\theta|I_{k-1})}} +  p(\theta|I_{N-1})\ln{\frac{p(\theta|I_0)}{p(\theta|I_{N-1})}} \]\,d\theta\\
    &=\int_{\Theta}\EE_{I_1,\dots,I_{N-1}|\pi,x_0} \[ \sum_{k=1}^{N-2} p(\theta|I_k)\ln{\frac{p(\theta|I_k)}{p(\theta|I_{k-1})}} +  p(\theta|I_{N-1})\ln{\frac{p(\theta|I_0)}{p(\theta|I_{N-2})}} \]\,d\theta\\
    &=\int_{\Theta}\EE_{I_1,\dots,I_{N-2}|\pi,x_0} \[ \sum_{k=1}^{N-3} p(\theta|I_k)\ln{\frac{p(\theta|I_k)}{p(\theta|I_{k-1})}} +  p(\theta|I_{N-2})\ln{\frac{p(\theta|I_0)}{p(\theta|I_{N-3})}} \]\,d\theta\\
    &\qquad \vdots \\
    &=\int_{\Theta}\EE_{I_1|\pi,x_0} \[ p(\theta|I_{1})\ln{\frac{p(\theta|I_0)}{p(\theta|I_0)}} \]\,d\theta\\
    &=0,
\end{align*}
\\\\\\\\\\\\\\\\\\\\\\where the third equality takes the last term from the sigma-summation and combines it with the last term, the fourth equality expands the expectation and uses $p(I_N|I_1,\ldots,I_{N-1},\pi) = p(I_N|I_{N-1},\pi)$, the fifth equality makes use of $p(\theta|I_N)=p(\theta|I_N,\pi)$, and the seventh to second-to-last equalities repeat the same procedures recursively. 
Hence, $U_T(\pi)=U_I(\pi)$.
\end{proof}

\section{Policy Gradient Expression}
\label{app:pg_derive}

Our proof for \cref{thm:PG} follows the proof given by \cite{silver2014deterministic} for a general infinite-horizon  MDP. 
Before presenting our proof, we first introduce a shorthand notation for writing the state transition probability:
\begin{align}
p(x_k \rightarrow x_{k+1}|\pi_w)=p(x_{k+1}|x_k,\mu_{k,w}(x_k)).
\end{align}
When taking an expectation over consecutive state transitions, we further use the simplifying notation
\begin{align}
&\int_{x_{k+1}}p(x_k \rightarrow x_{k+1}|\pi_w) \int_{x_{k+2}} p(x_{k+1} \rightarrow x_{k+2}|\pi_w) \nonumber\\
&\qquad \cdots \int_{x_{k+m}} p(x_{k+(m-1)} \rightarrow x_{k+m}|\pi_w) \[\cdots\] \,dx_{k+1} \, dx_{k+2} \cdots \, dx_{k+m} \nonumber\\ 
&= \int_{x_{k+m}} p(x_k \rightarrow x_{k+m}|\pi_w) \[\cdots\]  \, dx_{k+m}
\\ 
&= \EE_{x_{k+m} | \pi_w, x_k} \[\cdots\].
\end{align}

To avoid notation congestion, below we will also omit the subscript on $w$ and shorten  $\mu_{k,w_k}(x_k)$ to $\mu_{k,w}(x_k)$, with the understanding that $w$ takes the same subscript as the $\mu$ function. 

\begin{proof}[Proof of \cref{thm:PG}]

We begin by recognizing that the gradient of expected utility in \cref{eq:expected_utility_w} can be written using the V-function:
\begin{align}
    \nabla_w U(w) = \nabla_w V^{\pi_w}_0(x_0).\label{e:gradU_derive}
\end{align}
The goal is then to derive the gradient expression for the V-functions. 

We apply the definitions and recursive relations for the V- and Q-functions, and obtain a recursive relationship for the gradient of V-function:
\begin{align}
    \nabla_w V^{\pi_w}_k(x_k) 
    &= \nabla_w Q^{\pi_w}_k(x_k,\mu_{k,w}(x_k)) 
        \nonumber\\
    &= \nabla_w \Bigg[ \int_{y_k} p(y_k|x_k,\mu_{k,w}(x_k))g_k(x_k,\mu_{k,w}(x_k),y_k)\,dy_k \nonumber\\
    &\qquad\qquad + \int_{x_{k+1}} p(x_{k+1}|x_k,\mu_{k,w}(x_k)) V^{\pi_w}_{k+1}(x_{k+1}) \,dx_{k+1} \Bigg] \nonumber\\
    &= \nabla_w \int_{y_k} p(y_k|x_k,\mu_{k,w}(x_k))g_k(x_k,\mu_{k,w}(x_k),y_k)\,dy_k \nonumber\\
    &\qquad\qquad + \nabla_w \int_{x_{k+1}} p(x_{k+1}|x_k,\mu_{k,w}(x_k)) V^{\pi_w}_{k+1}(x_{k+1}) \,dx_{k+1} \nonumber\\
    &= \int_{y_k} \nabla_w \mu_{k,w}(x_k) \nabla_{d_k} \[ p(y_k|x_k,d_k) g_k(x_k,d_k,y_k) \]\Big|_{d_k=\mu_{k,w}(x_k)} \,dy_k \nonumber\\
    &\qquad\qquad + \int_{x_{k+1}} \Big[ p(x_{k+1}|x_k,\mu_{k,w}(x_k)) \nabla_w V^{\pi_w}_{k+1}(x_{k+1}) 
    \nonumber\\ 
    &\qquad\qquad +\nabla_w \mu_{k,w}(x_k) \nabla_{d_k} p(x_{k+1}|x_k,d_k)\Big|_{d_k=\mu_{k,w}(x_k)} V^{\pi_w}_{k+1}(x_{k+1}) \Big] \,dx_{k+1} \nonumber\\
    &= \nabla_w \mu_{k,w}(x_k) \nabla_{d_k} \Bigg[ \int_{y_k} p(y_k|x_k,d_k) g_k(x_k,d_k,y_k) \,dy_k 
    \nonumber\\
    &\qquad\qquad\qquad\qquad + \int_{x_{k+1}}  p(x_{k+1}|x_k,d_k)V^{\pi_w}_{k+1}(x_{k+1})dx_{k+1} \Bigg]\Bigg\vert_{d_k=\mu_{k,w}(x_k)} \nonumber\\
    &\qquad\qquad + \int_{x_{k+1}} p(x_{k+1}|x_k,\mu_{k,w}(x_k)) \nabla_w V^{\pi_w}_{k+1}(x_{k+1}) \,dx_{k+1} \nonumber\\
    &= \nabla_w \mu_{k,w}(x_k) \nabla_{d_k} Q^{\pi_w}_{k}(x_k,d_k)\Big|_{d_k=\mu_{k,w}(x_k)} 
    \label{e:gradV_recursive}\\
    &\qquad\qquad + \int_{x_{k+1}} p(x_k \rightarrow x_{k+1}|\pi_w) \nabla_w V^{\pi_w}_{k+1}(x_{k+1}) \,dx_{k+1}. \nonumber
\end{align}
Applying the recursive formula \cref{e:gradV_recursive} to itself repeatedly and expanding out the overall expression,
we obtain
\begin{align}
    &\nabla_w V^{\pi_w}_k(x_k) \nonumber\\
    &= \nabla_w \mu_{k,w}(x_k) \nabla_{d_k} Q^{\pi_w}_k(x_k,d_k)\Big|_{d_k=\mu_{k,w}(x_k)} \nonumber\\
    &\qquad + \int_{x_{k+1}} p(x_k \rightarrow x_{k+1}|\pi_w) \nabla_w \mu_{k+1,w}(x_{k+1}) \nabla_{d_{k+1}} Q^{\pi_w}_{k+1}(x_{k+1},d_{k+1})\Big|_{d_{k+1}=\mu_{k+1,w}(x_{k+1})} \,dx_{k+1} \nonumber\\
    &\qquad + \int_{x_{k+1}} p(x_k \rightarrow x_{k+1}|\pi_w) \int_{x_{k+2}} p(x_{k+1} \rightarrow x_{k+2}|\pi_w) \nabla_w V^{\pi_w}_{k+2}(x_{k+2}) \,dx_{k+2} \,dx_{k+1} \nonumber\\
    &= \nabla_w \mu_{k,w}(x_k) \nabla_{d_k} Q^{\pi_w}_k(x_k,d_k)\Big|_{d_k=\mu_{k,w}(x_k)} \nonumber\\
    &\qquad + \int_{x_{k+1}} p(x_k \rightarrow x_{k+1}|\pi_w) \nabla_w \mu_{k+1,w}(x_{k+1}) \nabla_{d_{k+1}} Q^{\pi_w}_{k+1}(x_{k+1},d_{k+1})\Big|_{d_{k+1}=\mu_{k+1,w}(x_{k+1})} \,dx_{k+1} \nonumber\\
    &\qquad + \int_{x_{k+2}} p(x_{k} \rightarrow x_{k+2}|\pi_w) \nabla_w V^{\pi_w}_{k+2}(x_{k+2}) \,dx_{k+2} \nonumber\\
    &= \nabla_w \mu_{k,w}(x_k) \nabla_{d_k} Q^{\pi_w}_k(x_k,d_k)\Big|_{d_k=\mu_{k,w}(x_k)} \nonumber\\
    &\qquad + \int_{x_{k+1}} p(x_k \rightarrow x_{k+1}|\pi_w) \nabla_w \mu_{k+1,w}(x_{k+1}) \nabla_{d_{k+1}} Q^{\pi_w}_{k+1}(x_{k+1},d_{k+1})\Big|_{d_{k+1}=\mu_{k+1,w}(x_{k+1})} \,dx_{k+1} \nonumber\\
    &\qquad + \int_{x_{k+2}} p(x_k \rightarrow x_{k+2}|\pi_w) \nabla_w \mu_{k+2,w}(x_{k+2}) \nabla_{d_{k+2}} Q^{\pi_w}_{k+2}(x_{k+2},d_{k+2})\Big|_{d_{k+2}=\mu_{k+2,w}(x_{k+2})} \,dx_{k+2}\nonumber\\
    &\hspace{2.5em}\vdots \nonumber\\
    &\qquad + \int_{x_{N}} p(x_{k} \rightarrow x_{N}|\pi_w) \nabla_w V^{\pi_w}_{N}(x_{N}) \,dx_{N} \nonumber\\
    &= \sum_{l=k}^{N-1} \int_{x_l} p(x_k \rightarrow x_l|\pi_w) \nabla_w \mu_{l,w}(x_l) \nabla_{d_l} Q^{\pi_w}_l(x_l,d_l)\Big|_{d_l=\mu_{l,w}(x_l)} \,dx_l\nonumber\\
    &= \sum_{l=k}^{N-1} \EE_{x_l| \pi_w, x_k} \[\nabla_w \mu_{l,w}(x_l) \nabla_{d_l} Q^{\pi_w}_l(x_l,d_l)\Big|_{d_l=\mu_{l,w}(x_l)}\] \,dx_l,\label{e:gradV_final}
\end{align}
where for the second-to-last equality, we absorb the first term into the sigma-notation by using
\begin{align*}
& \nabla_w \mu_{{k},w}(x_{k}) \nabla_{d_{k}} Q^{\pi_w}_{k}(x_{k},d_{k})\Big|_{d_{k}=\mu_{k,w}(x_{k})} \nonumber\\
& \qquad = \int_{x_{k}} p(x_k | x_{k},\mu_{k,w}(x_k)) \nabla_w \mu_{{k},w}(x_{k}) \nabla_{d_{k}} Q^{\pi_w}_{k}(x_{k},d_{k})\Big|_{d_{k}=\mu_{k,w}(x_{k})} \,dx_{k}
\nonumber\\
& \qquad = \int_{x_{k}} p(x_k \rightarrow x_{k}|\pi_w) \nabla_w \mu_{{k},w}(x_{k}) \nabla_{d_{k}} Q^{\pi_w}_{k}(x_{k},d_{k})\Big|_{d_{k}=\mu_{k,w}(x_{k})} \,dx_{k},
\end{align*}
and we eliminate the last term in the summation since
$\nabla_w V^{\pi_w}_{N}(x_{N})=\nabla_w g_{N}(x_{N})=0$.

At last, substituting \cref{e:gradV_final} into 
\cref{e:gradU_derive}, we obtain the policy gradient expression:
\begin{align}
    \nabla_w U(w) &= \nabla_w V^{\pi_w}_0(x_0) \nonumber\\
    &= \sum_{l=0}^{N-1} \EE_{x_l|\pi_w,x_0} \[ \nabla_w \mu_{l,w}(x_l) \nabla_{d_l} Q^{\pi_w}_l(x_l,d_l)\Big|_{d_l=\mu_{l,w}(x_l)} \]. \nonumber
    \end{align}
Renaming the iterator from $l$ to $k$ arrives at \cref{eq:pg_theorem} in \cref{thm:PG}, completing the proof.
\end{proof}

\section*{Acknowledgments}

This research is based upon work supported in part by the U.S. Department of Energy, Office of Science, Office of Advanced Scientific Computing Research, under Award Number DE-SC0021398. This paper was prepared as an account of work sponsored by an agency of the United States Government. Neither the United States Government nor any agency thereof, nor any of their employees, makes any warranty, express or implied, or assumes any legal liability or responsibility for the accuracy, completeness, or usefulness of any information, apparatus, product, or process disclosed, or represents that its use would not infringe privately owned rights. Reference herein to any specific commercial product, process, or service by trade name, trademark, manufacturer, or otherwise does not necessarily constitute or imply its endorsement, recommendation, or favoring by the United States Government or any agency thereof. The views and opinions of authors expressed herein do not necessarily state or reflect those of the United States Government or any agency thereof.

\bibliography{references}
\bibliographystyle{siamplain}

\end{document}